\documentclass[10pt,journal,compsoc]{IEEEtran}

\usepackage[ruled]{algorithm2e}
\usepackage{amsmath}
\usepackage{graphicx}
\usepackage{amssymb}
\usepackage{booktabs}
\usepackage{algorithmic}
\usepackage{color}
\usepackage{float}
\usepackage{wrapfig}
\usepackage{url}
\usepackage{array}
\usepackage{bm}

\ifCLASSOPTIONcompsoc
  \usepackage[nocompress]{cite}
\else
  \usepackage{cite}
\fi
\ifCLASSINFOpdf
\else
\fi

\ifCLASSOPTIONcompsoc
 \usepackage[caption=false,font=footnotesize,labelfont=sf,textfont=sf]{subfig}
\else
 \usepackage[caption=false,font=footnotesize]{subfig}
\fi

\begin{document}

\title{Pattern Guided UV Recovery for Realistic Video Garment Texturing }

\author{Youyi~Zhan,
        Tuanfeng~Y.~Wang,
        Tianjia~Shao,
        and~Kun~Zhou,~\IEEEmembership{Fellow,~IEEE}
        
\IEEEcompsocitemizethanks{\IEEEcompsocthanksitem Youyi Zhan, Tianjia Shao are with the State
Key Lab of CAD \& CG, Zhejiang University, Hangzhou 310058,
China. Tianjia Shao is the corresponding author of the work. E-mail: zhanyy@zju.edu.cn, tjshao@zju.edu.cn. 
\IEEEcompsocthanksitem Tuanfeng Y. Wang is with Adobe Research. E-mail: yangtwan@adobe.com.
\IEEEcompsocthanksitem Kun Zhou is with the State Key Lab of CAD \& CG, Zhejiang University,
Hangzhou 310058, China, and also with the FaceUnity Joint Lab of Intelligent Graphics, Zhejiang University, Hangzhou 310058, China.
\protect\\ E-mail: kunzhou@acm.org.}
\thanks{Manuscript received xxx; revised xxxx.}}

\markboth{IEEE TRANSACTIONS ON VISUALIZATION AND COMPUTER GRAPHICS}%
{Shell \MakeLowercase{\textit{et al.}}: Bare Demo of IEEEtran.cls for Computer Society Journals}

\IEEEtitleabstractindextext{%

\begin{abstract}
The fast growth of E-Commerce creates a global market worth USD 821 billion for online fashion shopping. What unique about fashion presentation is that, the same design can usually be offered with different cloths textures. However, only real video capturing or manual per-frame editing can be used for virtual showcase on the same design with different textures, both of which are heavily labor intensive. In this paper, we present a pattern-based approach for UV and shading recovery from a captured real video so that the garment's texture can be replaced automatically. The core of our approach is a per-pixel UV regression module via blended-weight multilayer perceptrons (MLPs) driven by the detected discrete correspondences from the cloth pattern. We propose a novel loss on the Jacobian of the UV mapping to create pleasant seams around the folding areas and the boundary of occluded regions while avoiding UV distortion. We also adopts the temporal constraint to ensure consistency and accuracy in UV prediction across adjacent frames. We show that our approach is robust to a variety type of clothes, in the wild illuminations and with challenging motions. We show plausible texture replacement results in our experiment, in which the folding and overlapping of the garment can be greatly preserved. We also show clear qualitative and quantitative improvement compared to the baselines as well. With the one-click setup, we look forward to our approach contributing to the growth of fashion E-commerce. 
\end{abstract}

}

\maketitle

\IEEEdisplaynontitleabstractindextext

\IEEEpeerreviewmaketitle


\IEEEraisesectionheading{\section{Introduction}}
\label{sec:intro}

\IEEEPARstart{T}{he} fast growth of E-Commerce creates a global market worth USD 821 billion for online fashion shopping in 2023 with a compound annual growth rate (CAGR) of $10.3\%$ \cite{fashion23}. The huge yet dynamic market accelerates the need for an efficient approach to present an apparel product virtually from the seller to the end customers. What unique about fashion E-commerce is that, a product (e.g., a shirt) is commonly designed with a fixed sewing but the fabric texture has a variety of optional patterns, i.e., the same shirt can be made of cornflower blue or maroon red, as shown in Figure~\ref{Fig:motivation}. Furthermore, it is imperative to experiment with various appearances in order to deliver the optimal outcome to customers in a fashion product advertisement. To present all these fabric textures choices to the customer, in the current industry workflow, the sellers either take photos/videos for all fabrics with high cost of hiring photo studios and fashion models, or seek professional photo editing services to replace the texture manually in Photoshop and AfterEffect as a post process, both of which are highly labor intensive. Hence, the market is looking for a better way to perform this task with lower cost and less manual intervention. 

Recent progress on dense UV estimation from portrait photos or videos~\cite{guler2018densepose,tan2021humangps, neverova2020continuous, xie2022temporaluv, ianina2022bodymap,jafarian2023normal} may shine a light on this automatic texturing task. Starting from a human portrait image, it calculates a mapping for the UV coordinate for each pixel. Such a method can be used for looking up albedo value from the texture map for each pixel on the image. However, the complexity of garment surface geometry together with self-occlusion makes such approach only produce a prediction of coarse UV, which makes it difficult to generate a plausible texture in the next step. These methods are also limited to the tight body or a specific type of clothes, which largely narrows the scenario in fashion creation. 
In order to improve the accuracy of UV estimation, another line of research proposes pattern based solutions~\cite{scholz2005garment, halimi2022garment}. They propose a printable pattern to facilitate more accurate correspondence estimation over the garment surface and use it for 3D geometry capturing. However, such route is mainly designed for 3D garment reconstruction by garment template fitting and is not ideal for the image based garment texturing application, because the fitted 3D mesh has limited capability of fully recovering detailed wrinkles in the image domain. Moreover, the synchronized multi-view 4K camera capturing setup in the lab environment is not suitable for ordinary users, who mostly use smartphones to capture single-view videos in wild conditions. Motion blur and different illuminations are common in these videos, and the correspondence estimation can fail in such conditions (see Figure~\ref{Fig:3mm_failure} for example).

\begin{figure*}[t]
    \begin{center}
    \includegraphics[width= 0.975\linewidth]{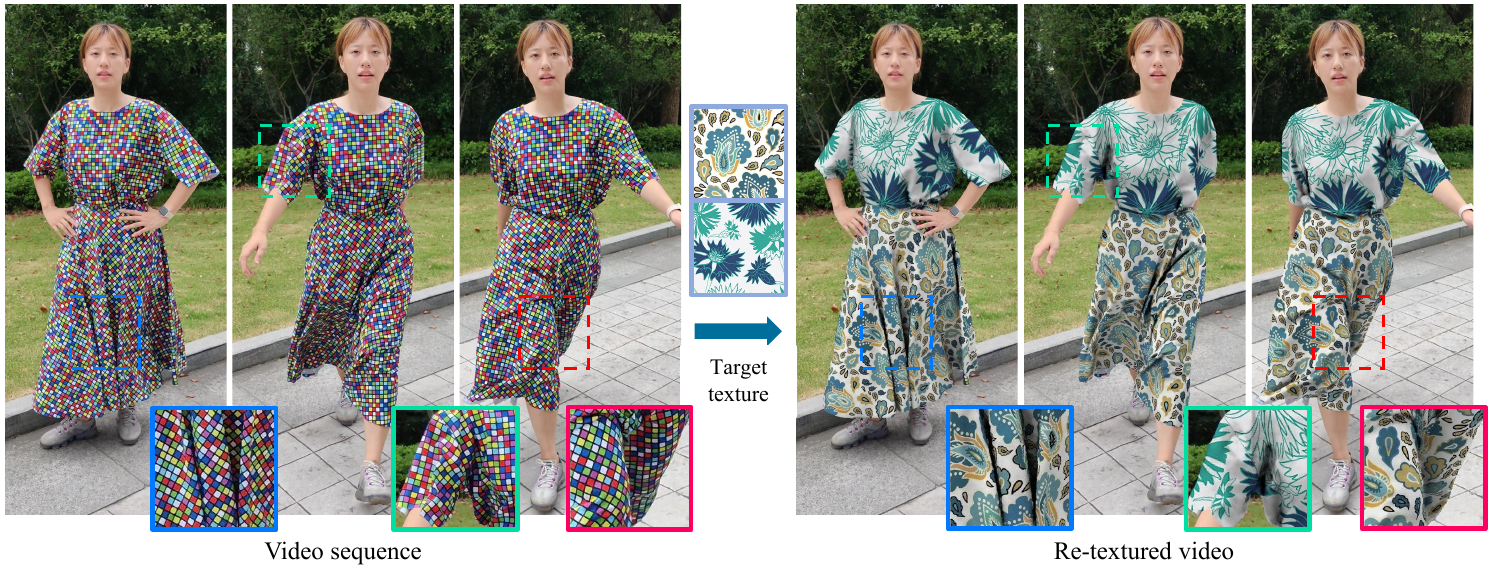}
    \end{center}
    \vspace{-5mm}
    \caption{We present an approach for pattern guided garment texture replacement in product virtual showcase. Given a video of a person wearing the target garments fabricated with our designed patterns, our method automatically extracts per-pixel UV coordinates for the garments in each frame as well as the shading and mask layers so that different texture maps can be applied to the image space for fast realistic visualization with just one click. The quality of our results lives up to the requirements from commercial applications where challenging areas such as folding areas and seams are nicely handled.}
    \label{Fig:teaser}
\end{figure*}

\begin{figure}[t]
  \begin{center}
    \includegraphics[width=0.475\textwidth]{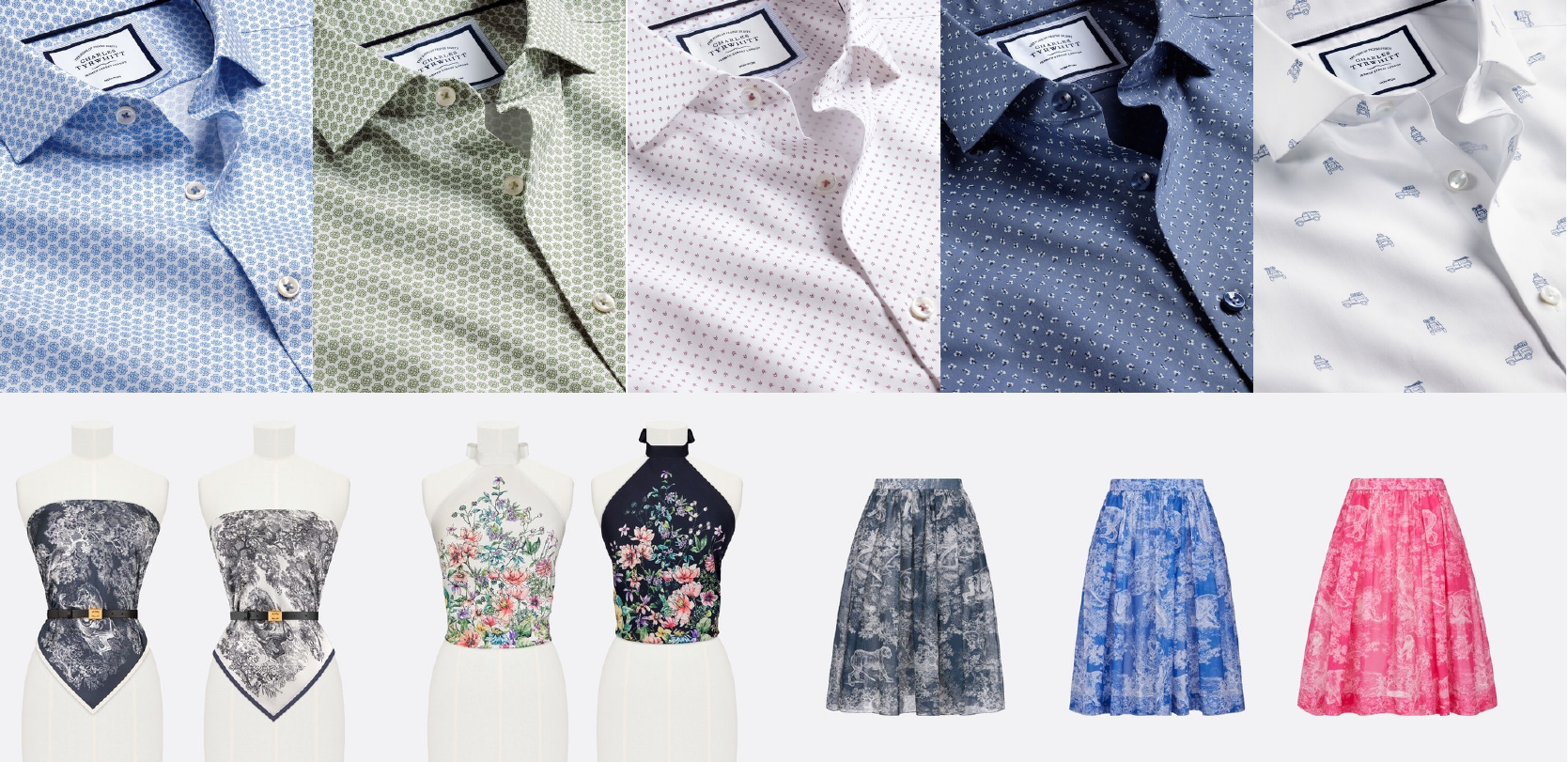}
  \end{center}
  \vspace{-5mm}
  \caption{What unique about fashion E-commerce is that, a product is commonly designed once but fabricated with a variety of optional patterns, i.e., the same shirt can be made of cornflower blue or maroon red. Image source: Dior\textcopyright~\cite{dior}, Charles Tyrwhitt\textcopyright~\cite{charlestyrwhitt}.}
  \label{Fig:motivation}
  \vspace{-5mm}
\end{figure}

In this paper, we seek a pattern-based approach for UV and shading reconstruction from a single-view video so that the garment can be realistically textured automatically. We use the pattern in~\cite{halimi2022garment}, but down-sample the grid resolution from 2.7mm to 15mm so that it can be clearly captured by a smartphone. There are three challenges for our task. First, the accuracy of correspondence estimation for the multi-view setup is not enough for single-view videos, which will cause wrong UV mapping in subsequent steps (see Figure~\ref{Fig:corres_improve} for example). Second, the estimated dense correspondences are still relatively sparse in the image space. We need to recover the UVs for all pixels while preserving complex geometry details (e.g., the foldings and occlusions). Third, the estimated UV mapping should be time-coherent to obtain consistent visual effects in the video.

Our method starts from the garment fabricated with the given pattern. To tackle the above challenges, for each video frame, we first propose a simple but effective graph-growing algorithm to improve the correspondence detection accuracy. We also propose a voting strategy to make it applicable to a person wearing multiple garments. Based on the detected discrete correspondences, we propose a novel multilayer perceptron (MLP) model to regress per-pixel UV coordinates in image space while preserving details for each frame. As the correspondences can still be missing in difficult areas (i.e., boundary areas around foldings and wrinkles), we constrain the Jacobian of the UV mapping to the predicted UV gradient to avoid distorted textures. To deal with the UV discontinuity caused by the folding and seam lines, we further propose a robust loss function to encourage the Jacobain outlier values to appear less often but more significant, which will make the outlier pixels assemble around the foldings and seams, producing results with visually plausible discontinuity details.

For the video input, the temporal consistency is the key for a better visual quality. Inspired by previous work~\cite{holden_phase-functioned_2017,dehesa_grid-functioned_2021,hao2022implicit}, we propose a blended weight model to ensure the consistency along temporal axis and avoid jittering. Large movements in the video also introduce motion blur and more occlusions, make it impossible to detect the correspondences for most of the areas. We combine the off-the-shelf optical flow estimation~\cite{teed2020raft, jiang2021learning} with the image inpainting technique~\cite{xu2019deep,gao2020flow} and link the nearby frames to compensate for the missing correspondences and estimate a reasonable and consistent UV map. Afterwards, we predict the shading from the input image via a U-Net model. Finally, each frame of the video can be re-textured using our predicted UV mapping and shading in a consistent manner.

We quantitatively and qualitatively evaluate our approach with synthetic datasets as well as real captured videos of clothes made by our designed patterns. We demonstrate clear improvements from the alternative solutions. We show that our approach is robust to complicated clothes - different types of fashion designs, in the wild illuminations, with challenging poses, body motions and body-garment interactions. With the simple yet fully automated setup, our approach shows great potential in product virtual showcase.

In summary, our contributions are as follows: 
1) a robust yet fully automated workflow for texture replacement on videos of the garments made by designed patterns;
2) a correspondence guided UV regression model capable of capturing discontinuity and avoiding distortion in UV coordinates;
3) a blended weight model and temporal constraint helping ensure the accuracy and consistency of the UVs in videos;
4) a simple correspondence detection algorithm on the designed pattern which provide $\sim20\%$ improvement compared to the SOTA solution.


\section{Related Work}
\label{sec:relatedwork}

\subsection{Human Body UV Prediction}

Establishing dense correspondences between a single human portrait image and a canonical parameterization space, a.k.a, UV prediction, has been proved critical for many downstream tasks, e.g., body tracking~\cite{tancik2020fourfeat,yan2021ultrapose}, and human reposing~\cite{sarkar2021humangan, albahar2021pose}. As one of the most successful solution so far, Densepose~\cite{guler2018densepose} offers a prediction of UV map of a tight human body from an input image. Learning a 2D UV embedding of a pre-defined atlas without any constraint on the target surface geometry may lead to strong artifacts. ~\cite{neverova2020continuous} addresses this issue by learning a latent parameterization space without being constrained on specific geometry types. HumanGPS~\cite{tan2021humangps} further incorporates geometric constraints by regularizing the UV prediction with geodesic distance. Due to the underlying tight body model and lack of training data, these approaches only run for people wearing tight clothes. Powered by a Vision Transformer, BodyMap~\cite{ianina2022bodymap} produces continuous correspondences for every foreground pixel of clothed human including loose clothes, hair, and accessories. Although the correspondence is as dense as per-pixel, the accuracy still falls short for the garment texturing task on the captured image. Besides, no local geometric cue from the image is used by these approaches. Therefore, the output UV prediction could hardly contain fine details, e.g., wrinkles. Jafarian et al.~\cite{jafarian2023normal} could capture the geometry details based on the predicted garment normal and obtain visually pleasant re-textured results. However, it cannot handle the folding and self-occlusion of the garment. In contrast to the above works, our approach based on printed pattern incorporates rich geometric details with occlusions correctly presented.

\subsection{Learning UV with Temporal Consistency}
As a natural extension for image based approaches, predicting UV for consecutive frames is actively explored in recent years. 
NeuralAtlas ~\cite{kasten2021layered} learns a set of unified representation of the appearance for each object from the input video by analyzing the optical flow across the frames. The learned 2D representation shared by all the frames could be directly used as a texture map for the object of interest. Deformable Sprites~\cite{ye2022sprites} further allows learning a dynamic representation for each layer of the video in an unsupervised manner. Such an approach does not assume humanoid shape, so no inter-subject correspondence could be established. Most recently, TemporalUV~\cite{xie2022temporaluv} expands Densepose~\cite{guler2018densepose} UV prediction to include everywhere within the body outlines and aims for temporal stability via a feature matching step. As heavily inherited from Densepose~\cite{guler2018densepose}, fine details are rarely captured by this approach. We adopt the idea of temporal consistency from such a method and achieve plausible performance on sequential input.

\subsection{Modeling Garment in 3D with Patterns}
The 3D mesh reconstruction for the target clothes may lead to easy UV parameterization afterwards. However, accurate 3D geometry recovery can be challenging and difficult to be reliable for our texture replacement task. Using color-coded pattern to guide 3D clothes capture is initially discussed by ~\cite{guskov2003trackable, scholz2005garment} and proven to be a success, since the pre-printed pattern helps the correspondence detection. Other technologies adopt invisible markers, e.g., infrared ink~\cite{narita2016dynamic}, for high quality dense correspondence mapping. However, the requirement for specialized fabric material and capture device makes such technology difficult to scale. GarmentAvatar~\cite{halimi2022garment} adopts the color pattern method and enhances the pipeline by plugging in the recent progress of using neural network for feature detection and then deforms a 3D garment template to reflect the details captured by the image features. We use the same pattern design from GarmentAvatar~\cite{halimi2022garment} but there are three main differences: 1) we do not use a 3D mesh to bridge the image to UV space, so the level of details is not bounded by the template mesh resolution; 2) Our method is not trained in a garment-specific manner, i.e., no specific dataset is required and can be directly applied to an arbitrary type of garment; 3) Our method is light-weight, working on a video captured by only one camera, so it can be used for wider scenarios. 


\section{Method}
\label{sec:method}
\noindent\textbf{Overview}
Our workflow starts from a video capturing a person wearing garments sewed from a fabric painted with the specifically designed texture pattern. We first introduce the algorithm for 2D pattern registration with our improved correspondence detection algorithm (Section~\ref{sec:corr}). The core of our approach is to establish per-pixel correspondences from each video frame to a shared UV parameterization space. We formulate this problem as a regression task and adopt multilayer perceptron (MLP) to solve it. For a single frame of the video, we use the detected correspondences as constraints to regress the MLP and introduce a novel loss term on the Jacobian of the predicted UVs to produce plausible UV predictions around the non-smooth areas caused by occlusions and seam cuts (Section~\ref{sec:dense}). When applying the method to a sequential input, i.e., a video footage, we adopt the blended weight technique and temporal loss (Section~\ref{sec:temp}) to improve the temporal coherence. Following the classical image composition workflow, we generate the albedo layer via the learned UV coordinates on each frame and learn the shading layer directly from the input image (Section~\ref{sec:shading}). The final image is generated with a given texture based on the predicted UVs and shading via a standard image rendering workflow. 

\begin{figure*}[t]
  \begin{center}
    \includegraphics[width=0.975\textwidth]{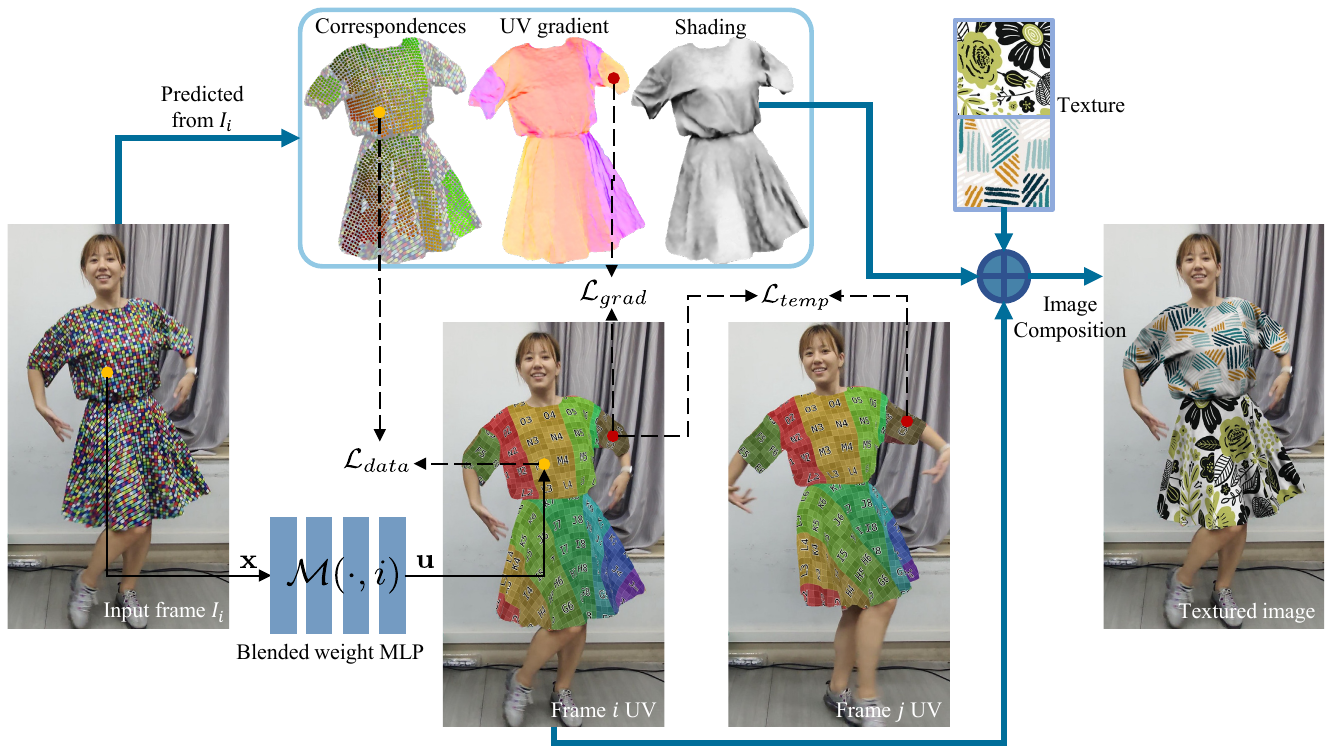}
  \end{center}
  \vspace{-5mm}
  \caption{Pipeline overview. Starting from an input frame $I_i$, we perform correspondence detection (Section~\ref{sec:corr}) to establish discrete correspondences between the image frame and the template pattern. Input coordinates are fed into the blended weight MLP (Section~\ref{sec:temp}) to regress the per-pixel UV coordinates, with both data constraint and gradient constraint (Section~\ref{sec:dense}). We also apply temporal constraints to neighboring frames to ensure the visual consistency (Section~\ref{sec:temp}). The shading and mask are also predicted from the input (Section~\ref{sec:shading}). The predicted UVs are used to texture the albedo layer with given texture maps, and the final images can be composed accordingly.
  }
  \label{Fig:pipeline}
\end{figure*}

\subsection{Pattern Guided Correspondence Detection}
\label{sec:corr}

To establish pixel-level correspondences between the image domain and UV space, the first step is to build discrete correspondences between them. We follow the color-coded patterns as in GarmentAvatar~\cite{halimi2022garment} for this task. We adopt a lower resolution pattern whose grid size is 15mm. The correspondence detection algorithm in~\cite{halimi2022garment} is designed for the setup of more than 200 camera views and a single garment, which are not suitable for our scenario, i.e., a single-view video of a person wearing multiple garments. The algorithm can fail on the boundary areas nearly perpendicular to the view direction, as well as the areas with motion blur, and are not able to distinguish the same local patterns but on different garments. To this end, we make essential modifications on the correspondence detection algorithms.

\begin{figure}[h]
  \begin{center}
    \includegraphics[width=0.49\textwidth]{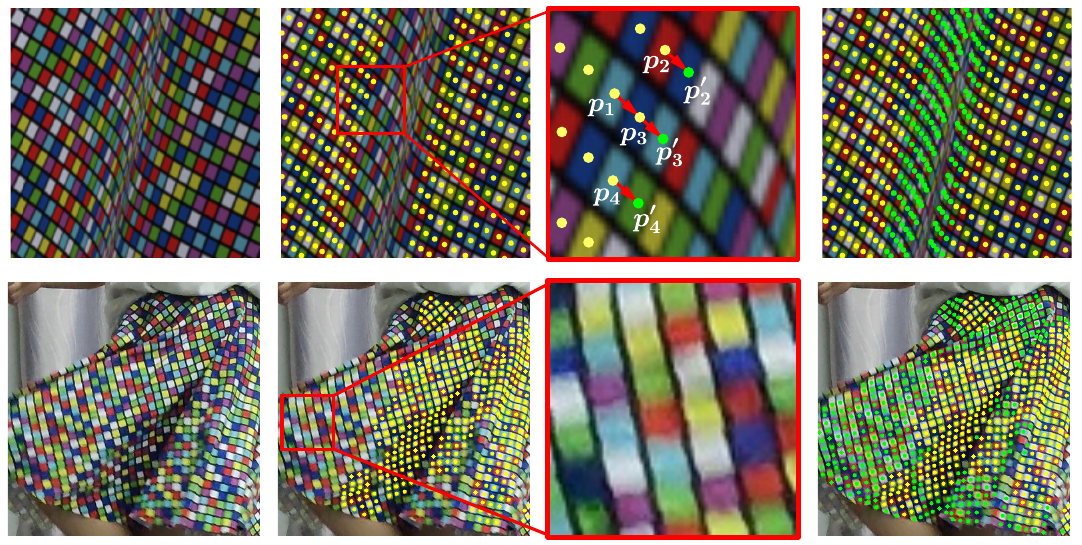}
  \end{center}
  \vspace{-5mm}
  \caption{Our correspondence detection enhances the results from GarmentAvatar~\cite{halimi2022garment} (yellow dots) and produces $19\%$ more correspondences (green dots). We present the boundary case (top row) and the motion blur case (bottom row) to show the efficacy of the improved detection algorithm.}
  \label{Fig:corr_refine}
\end{figure}

First, to make the the detection algorithm robust on the boundary and motion blur areas, we refine their detected homogeneous graph by growing the graph step by step. As shown in Figure~\ref{Fig:corr_refine} (top row), for a recognized area on the boundary, we move points $p_2,p_3,p_4$ with an offset $\overrightarrow{p_3-p_1}$ to get $p'_2,p'_3,p'_4$. With a distance threshold $\eta = \frac{1}{2}\|p_1-p_3\|$, we check if the center points detected near $p'_2,p'_3,p'_4$ as well as 
their colors agree with the expected pattern locations. By this refinement step, we can extend the correspondence set by $\sim19\%$ while keeping the precision drop negligible. The key observation of the above improvement is that the corner point and the homogeneous graph may not be correctly detected or constructed, but the center point as well as its color can be still detected and identified, which still provides enough information to recover the UV location of the center point. Please refer to our supplementary material for more details.

Second, to make the method applicable to multiple garments, we need more patterns for the fabrication of different garments. Although the $3\times 3$ grids are designed to be unique in one pattern, one $3\times 3$ grid can appear many times in multiple patterns, which may mislead our correspondence search. Therefore, we use a multiple pattern voting strategy to distinguish which pattern the correspondences belong to. Specifically, for a center point, we do neighbor voting in all the used patterns and get its major votes. The voting information contains both UV location and pattern id. We choose the pattern with the highest vote count as the center point’s correct pattern, and assign the pattern id and UV location to the center point. Please refer to our supplementary material for the algorithm.

\subsection{Single Frame UV Regression}
\label{sec:dense}
Once the correspondences are detected from the input single frame to the designed pattern, we use a multilayer perceptron (MLP) based regression network that predicts per-pixel UV coordinates over the image space, i.e.,
\begin{equation}
    \mathbf{u} = f(\mathbf{x} ; \bm{\theta}),
    \label{eqn:funcf}
\end{equation}
where $\mathbf{x} = (x,y) \in \mathbb{R}^2$ is the 2D pixel location in a frame. $\mathbf{u} = (u,v) \in \mathbb{R}^2$ is the corresponding UV coordinate. $f$ is the MLP function and $\bm{\theta}$ is the trainable parameters of the MLP. In the MLP, We also adopt Random Fourier Features~\cite{tancik2020fourfeat}, $\gamma(\mathbf{x}): \mathbb{R}^2 \mapsto \mathbb{R}^{256}$, to let our network learn high frequency coordinate signals in the 2D image domain. Based on the detected correspondences $\Sigma = \{\mathbf{x}_i \mapsto \mathbf{u}_i\}_{i=1,2,\cdots, N}$, we formulate the supervision from detected correspondences by a data loss, $\mathcal{L}_{data}$, as 
\begin{equation}
    \mathcal{L}_{data} = \sum_{\mathbf{x} \mapsto \mathbf{u} \in \Sigma} \|\mathbf{u} - f(\mathbf{x})\|.
\end{equation}

The detected correspondences, $\Sigma$, are still relatively sparse compared to the pixel resolution. This problem is even severer when occlusion happens (e.g., heavy wrinkles, overlapped layers, self-occlusion, etc.) as some of the correspondences near the boundaries are difficult to be established due to the high distortion and heavy shadow (see Figure~\ref{Fig:lgrad} for example). In such cases, the UV values may not be correctly regressed by the MLP with limited constraints. To address the problem, we observe that the grid-like pattern has a strong hint on the trend of the UV direction, and we could use the pattern gradient information retrieved directly from the image space to help to regress a more accurate UV.  We formulate it as the gradient loss, as
\begin{equation}
    \mathcal{L}_{grad} = \sum_{\mathbf{x} \in \Theta} \| \|\mathbf{J}_f(\mathbf{x}) - \mathbf{g}(\mathbf{x})\|_F \|_{0.5},
\end{equation}
where $\Theta$ are the areas where correspondences are missing (see Figure~\ref{Fig:lgrad} for the illustration). $\|\cdot \|_F$ is the Frobenius norm. $\mathbf{J}_f(\mathbf{x}) = \partial f(\mathbf{x}) / \partial \mathbf{x} : \mathbb{R}^2 \mapsto \mathbb{R}^{2\times 2}$ is the Jacobian of the output UV referring to the input pixel location. $\mathbf{g}: \mathbb{R}^2 \mapsto \mathbb{R}^{2\times 2}$ is the UV gradient predicted from the image space. Similar to image gradient~\cite{imagegradient}, we define the UV gradient $\mathbf{g}$ as the convolution result of the UV map with Sobel kernel~\cite{sobel}, at both x and y directions. Please refer to the supplementary material for the computation process of the UV gradient.

Please note that we apply $L_{0.5}$ loss for the gradient loss term. We follow the function format in \cite{barron2019general} and define $L_{0.5}$ loss as
\begin{equation}
L_{0.5}(x) = (x^2+c^4)^{0.25} - c,
\end{equation}
where $c=0.1$ is a smooth parameter controlling the loss near $x=0$. Given the fact that the MLP is $C^0$ continuous everywhere, the mapping from image to UV space implemented with MLP is guaranteed to be smooth. However, this is not appreciated when modeling the UV correspondences for the garment image. 
\setlength{\intextsep}{0pt}%
\setlength{\columnsep}{4pt}%
\begin{wrapfigure}{r}{0.3\textwidth}
  \centering\includegraphics[width=0.3\textwidth]{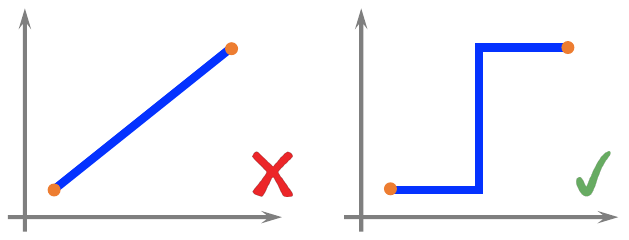}
\end{wrapfigure}
As shown in Figure~\ref{Fig:uv_discontinuity}, due to the occlusion, UV coordinates may have sudden change in neighbouring pixels of the image plane. Smooth transition between the neighbouring correspondences around the seams creates strong artifacts (see Figure~\ref{Fig:uv_discontinuity}). As shown in the inset figure, intuitively we ask the network not to average the loss all over the whole area but keep it to a specific region as small as possible. Compared with $L_2$ and $L_1$ losses, the $L_{0.5}$ loss suppresses the penalty caused by large errors~\cite{barron2019general}, which makes it tend to ignore the large errors around the seam areas where the gradients are difficult to be accurately predicted, and focus on other smooth areas on the garments. In this way, the $L_{0.5}$ loss helps regress the MLP jacobian $\mathbf{J}_f$ to the predicted UV gradient $\mathbf{g}$ for most areas but leaves the seams as the outliers, which produces sudden changes of UV and obtains better visual results.

The total loss used for UV reconstruction from a single frame is
\begin{equation}
    \mathcal{L}_{image} = \mathcal{L}_{data} + \lambda_{grad}\mathcal{L}_{grad},
\end{equation} where $\lambda_{grad}$ is the weight balancing the two terms.

\begin{figure}[t]
  \begin{center}
    \includegraphics[width=0.475\textwidth]{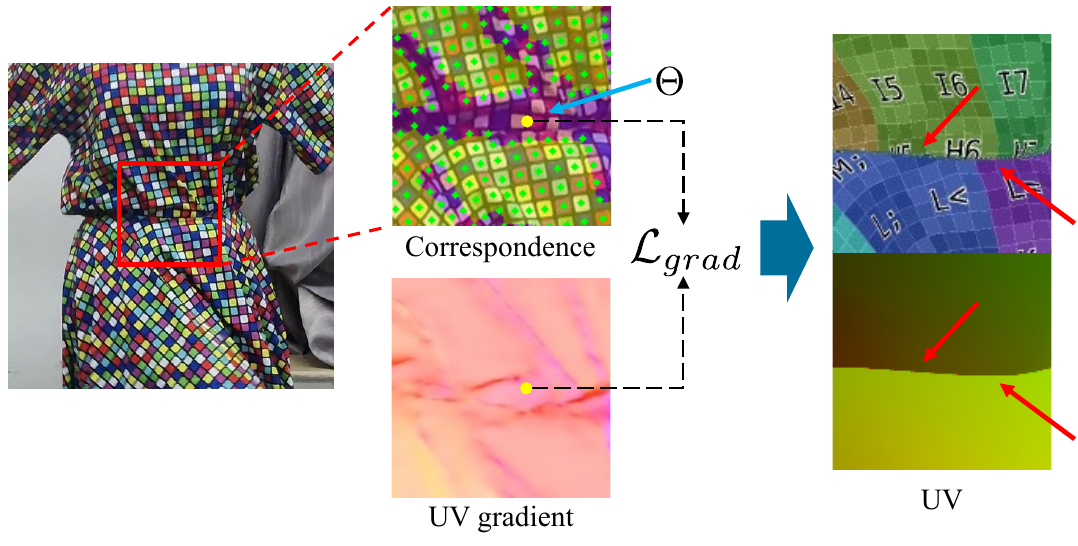}
  \end{center}
  \vspace{-5mm}
  \caption{The green dots represent the correspondences, and the regions not covered by yellow color are $\Theta$, which means correspondences are missing here. Correspondences could be difficult to be detected at $\Theta$ due to wrinkles and occlusions. We apply the $L_{0.5}$ gradient loss on $\Theta$ to constrain UV. Such constraint avoids UV smooth transition and produces discontinuity of UV at the folding and occlusion areas.}
  \label{Fig:lgrad}
\end{figure}

\subsection{UV Regression on Sequential Input}
\label{sec:temp}

\begin{figure}[t]
  \begin{center}
    \includegraphics[width=0.475\textwidth]{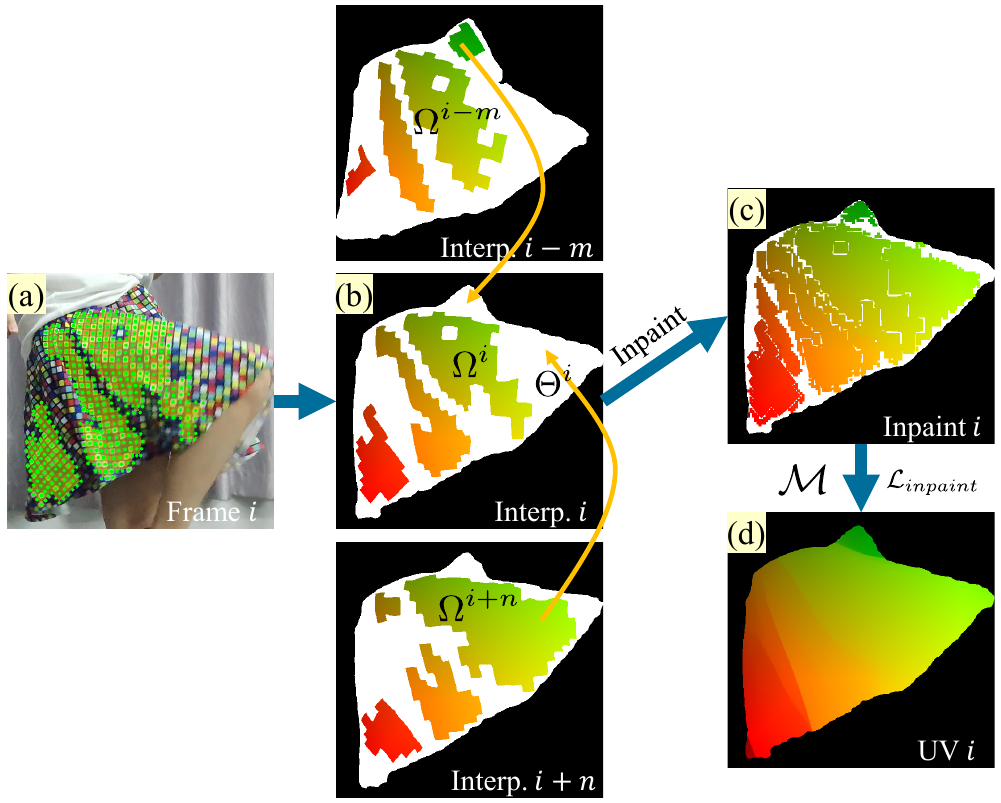}
  \end{center}
  \vspace{-5mm}
  \caption{ (a) The green dots represent the identified correspondences. The regions where these correspondences exist are represented by yellow color, which are also referred to as $\Omega^i$. (b) Within the domain $\Omega^i$, the values of UV can be interpolated using the correspondences based on the homogeneous graph. (c) Pixels in $\Theta^i$ can be inpainted by $\Omega$ of nearby frames. (d) By constraining $\mathcal{M}$ with the inpainted points, the UV of the whole garment could be recovered. } 
  \label{Fig:inpaint}
\end{figure}

When it comes to the sequential input from the video, the UV regression model can be formulated as follows,
\begin{equation}
   \mathbf{u} = \mathcal{M}(\mathbf{x},t),
\end{equation}
where $t= 1,\cdots, T$ is the time label of the sequential $T$ frames. For the frame at time $t$, the correspondences $\Sigma^t = \{\mathbf{x}_i^t \mapsto \mathbf{u}_i^t\}_{i=1,2,\cdots, N^t}$ can be detected as in Section~\ref{sec:corr}. To model $\mathcal{M}$, the easiest way is extending the single frame regression method (Section~\ref{sec:dense}) to the sequential input, i.e., predicting UV coordinates for all the frames in a video footage. However, directly applying the method independently to each frame will yield unsatisfactory results. The detected correspondences are relatively sparse for the image pixel space, and can vary between frames, resulting in jittering UV changes in adjacent frames and flickering artifacts. Moreover, large percentage of correspondences may not be detected at all in some frames due to severe motion blur (e.g., Figure~\ref{Fig:inpaint} and Figure~\ref{Fig:blur}). In these cases, we wouldn't obtain visually pleasant UV regression. We propose two strategies to solve the above problem, that is, the blended weight MLP and a temporal consistency term.

Input-dependent weights in neural network have been proven to have a better learning ability and smooth transitions as the input changes~\cite{holden_phase-functioned_2017,dehesa_grid-functioned_2021,hao2022implicit}. We adopt the same idea and blend the model weights (parameters) along the temporal axis. Specifically, instead of using a separate MLP network for each frame, we define a set of MLP parameters $\bm{\theta}_{\mathcal{M}} = \{\bm{\theta}_{i} | i=0,\cdots,M-1 \}$ for the whole video. $\bm{\theta}_{i}$ means the $i$-th weights defined at time $i*s$. Here $s$ is the time step, so for a video of length $T$, there are total $M = \lceil T/s \rceil + 1$ sets of MLP parameters. 
For a frame at time $t$, the corresponding MLP weights are computed as a linear blending of the parameter set $\bm{\theta}_{\mathcal{M}}$. Mathematically, $\mathcal{M}$ can be defined as
\begin{equation}
    \mathcal{M}(\mathbf{x},t) = f(\mathbf{x}; \sum_{i=0}^{M-1} \alpha_{it} \bm{\theta}_{i} ).
    \label{eqn:funcmodel}
\end{equation}
Here $\alpha_{it}$ is the blending coefficient calculated for time $t$. We use the cubic spline interpolation to compute $\alpha_{it}$ for the smooth change with respect to frame $t$, which is defined as follows. 

\begin{equation}
\begin{aligned}
\omega &= (t \pmod s)/s, \\
j &= \lfloor t/s \rfloor, \\
\alpha_{it} &= 
\begin{cases}
\omega, & i=j-1 \\
-\frac{1}{2} + \frac{1}{2} \omega^2, & i=j \\
1 - \frac{5}{2} \omega + 2 \omega^2 - \frac{1}{2} \omega^3, &i=j+1 \\
- \frac{1}{2} + \frac{3}{2} \omega - \frac{3}{2} \omega^2 + \frac{1}{2} \omega^3, &i=j+2 \\
0, & \text{others}.
\end{cases}
\end{aligned}
\label{eqn:cubicinter}
\end{equation}
\setlength{\intextsep}{0pt}%
\setlength{\columnsep}{4pt}%
\begin{wrapfigure}{r}{0.2\textwidth}
  \centering\includegraphics[width=0.2\textwidth]{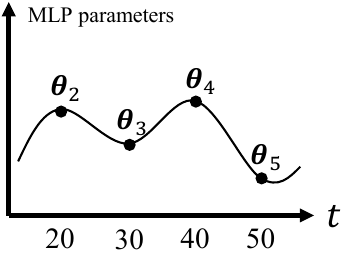}
\end{wrapfigure}
We also add extra padding MLP parameters $\bm\theta_{-1},\bm\theta_{M}$ to make the beginning and last several frames also work in Eq~\ref{eqn:cubicinter}. In practice, we set $s=10$ for a 30 fps video. The inset figure visualises the interpolation spline. The data loss and the gradient loss have the similar formulation under the blended weight MLP.
\begin{equation}
\begin{aligned}
\mathcal{L}_{data} &= \sum_{t=1}^{T} \sum_{\mathbf{x} \mapsto \mathbf{u} \in \Sigma^t} \|\mathbf{u} - \mathcal{M}(\mathbf{x},t)\|, \\
\mathcal{L}_{grad} &= \sum_{t=1}^{T} \sum_{\mathbf{x} \in \Theta^t} \| \|\mathbf{J}_{\mathcal{M}(\cdot,t)} (\mathbf{x}) - \mathbf{g}^t(\mathbf{x})\|_F \|_{0.5},
\end{aligned}
\end{equation}
where $\Theta^t$ is correspondence missing areas at time $t$, and $\mathbf{g}^t$ is the predicted UV gradient at time $t$. 

Using the blended weights can greatly reduce the jittering artifact in the predicted UVs. However, since the detected correspondences are relatively sparse in a frame and may vary in the consecutive frames, the consistency and accuracy of the UVs still cannot be ensured. To address this issue, we adopt the optical flow based temporal loss. We first predict the forward and backward optical flow between the adjacent frames, using the method in~\cite{teed2020raft}. The flows establish a connection between points in adjacent frames. We also conduct forward-backward consistency check to filter out the invalid point pairs. 
Based on the above point connections, our temporal loss is composed of two parts: an inpainting loss and a consistency loss. 

For the inpainting loss, since the missing correspondences at frame $i$ may appear in nearby frames, inspired by the video completion works~\cite{gao2020flow,xu2019deep}, we complete the missing areas at frame $i$ by inpainting the UV values propagated from the nearby frames. We define the pixel set $\Theta$ and $\Omega$ as shown in Figure~\ref{Fig:inpaint}. $\Omega$ is the areas where the correspondences exist, while $\Theta$ is the areas where the correspondences are absent. Inside $\Omega$ we could build the homogeneous graph~\cite{halimi2022garment} and fill the UVs by interpolating the correspondences. For the pixel $\mathbf{x}^i \in \Theta^i$, we follow the forward and backward flow links until reaching pixel $\bar{\mathbf{x}}^j \in \Omega^j$ at frame $j$, and fill $\mathbf{x}^i$ with the interpolated UV ${\bar{\mathbf{u}}}^j$. If $\mathbf{x}^i \in \Theta^i$ is connected with both forward and backward pixels $\bar{\mathbf{x}}^j \in \Omega^j, \bar{\mathbf{x}}^k \in \Omega^k$, it will be filled by a linear combination of their pixel values whose weights are inversely proportional to their distances to $\mathbf{x}^i$~\cite{xu2019deep,okabe2019interactive},\

\begin{equation}
    \mathbf{u}(\mathbf{x}^i) = \frac{ \bar{\mathbf{u}}^j / |i-j| + \bar{\mathbf{u}}^k / |i-k| }{ 1/|i-j| + 1/|i-k| }.
\end{equation}
Figure~\ref{Fig:inpaint} shows the interpolation and inpainted results. The inpainting loss is formulated as
\begin{equation}
    \mathcal{L}_{inpaint} = \sum_{t=1}^{T} \sum_{\mathbf{x} \in \Theta^t} ||\mathcal{M}(\mathbf{x},t)-\mathbf{u}^t(\mathbf{x})||.
\end{equation}

For the consistency loss, similar to~\cite{kasten2021layered, ye2022deformable}, we constrain the UV values of two points linked by the flow in adjacent frames. We use the following loss term,
\begin{equation}
    \mathcal{L}_{consist} = \sum_{t=1}^{T} \sum_{\mathbf{x} \in S^t, \mathbf{x}' \in S^{t'}}       \|\mathcal{M}(\mathbf{x},t) - \mathcal{M}(\mathbf{x}',t') \|,
\end{equation}
where $S^t$ is the garment mask at frame $t$, $t' \in \{t \pm 1, t\pm 2, t\pm 3\}$, and $\mathbf{x}'$ is the pixel propagated from $\mathbf{x}$, from frame $t$ to frame $t'$, based on the computed optical flow.

Together we have the temporal loss and the total loss for the video as
\begin{equation}
\begin{aligned}
    \mathcal{L}_{temp} &= \mathcal{L}_{inpaint} + \mathcal{L}_{consist}, \\
     \mathcal{L}_{video} &= \mathcal{L}_{data} + \lambda_{grad}\mathcal{L}_{grad} + \lambda_{temp}\mathcal{L}_{temp},
\end{aligned}
     \label{eqn:videoloss}
\end{equation} where $\lambda_{temp}$ is the weight for the temporal loss term.

\subsection{Shading, Mask and UV Gradient Prediction}
\label{sec:shading}
\begin{figure}[t]
  \begin{center}
    \includegraphics[width=0.499\textwidth]{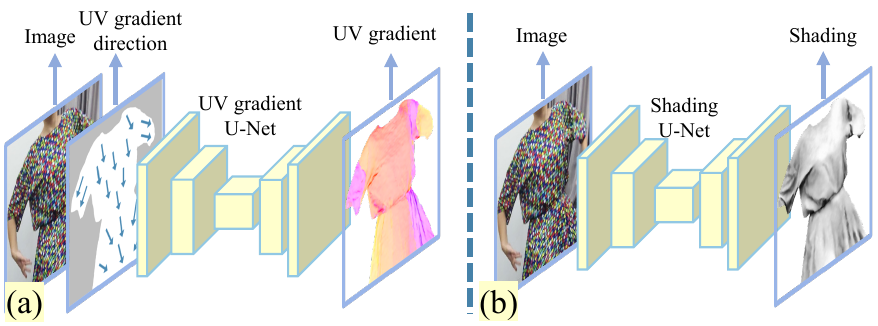}
  \end{center}
  \vspace{-5mm}
  \caption{ (a) Given the image and UV gradient direction as input, the UV gradient can be predicted by the UV gradient U-Net. (b) Given the image as input, the shading can be predicted by the shading U-Net. } 
  \label{Fig:unet}
\end{figure}

Following the workflow of intrinsic image decomposition~\cite{bell2014intrinsic}, we decompose the input image into the albedo and shading layer for the subsequent texture replacement. For this task, we train a network to predict the shading layer from the input image. Since the garment is captured with the pre-defined pattern, which implies the albedo layer is highly structured and predictable, a simple U-Net~\cite{ronneberger2015u} is powerful enough for this task. The shading U-Net accepts the image as input and predicts the shading layer (see Figure~\ref{Fig:unet} (b) for illustration).

Similarly, we use the same U-Net architecture to train a network to predict the UV gradient from our input image. However, prediction only from the input image results in ambiguity, as the UV gradient can have four possible directions on such grid-like patterns. We solve this by adding the UV gradient direction as a prior. During inference, we firstly regress the UV using Eq~\ref{eqn:videoloss}, with fewer training iterations and setting $\lambda_{grad}$ to zero. Then, we compute the UV gradient direction from the UV. Finally, we stack the UV gradient direction and the image, feed them into U-Net, and predict the UV gradient, as shown in Figure~\ref{Fig:unet} (a). Please refer to the supplementary material for the computation process of the UV gradient and gradient direction.

To segment the garments from each frame, we use Segment Anything~\cite{kirillov2023segment} to predict the masks. The detected correspondences can be used to help the model segment precise garment masks. Note that our method segments all the textured parts as a whole and does not need to split it into different garment pieces further.

With the shading and mask layer, we compose the final image as shown in Figure~\ref{Fig:pipeline}. We always apply \textit{Gamma correction} to the input image and inverse it back after composing the final output.


\begin{figure*}[t]
  \begin{center}
    \includegraphics[width=0.975\textwidth]{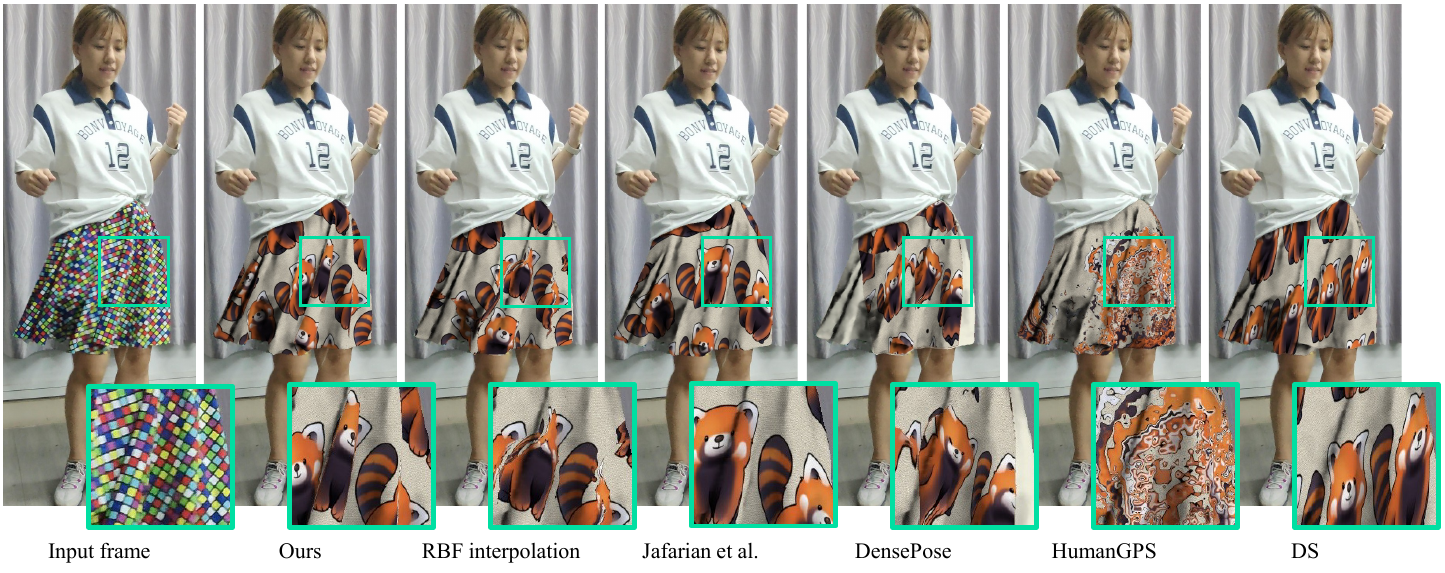}
  \end{center}
  \vspace{-5mm}
  \caption{Our approach outperforms the baselines in both UV accuracy and the texture-geometry alignment around the seams and wrinkles. }
  \label{Fig:comparison}
\end{figure*}

\section{Evaluation}
\label{sec:evaluation}
In this section, we evaluate our approach on multiple datasets and settings. We compare our approach with the state-of-the-art methods for dynamic garment UV recovery from synthetic or real video sequences. We also present a visual collection of our results (Figure~\ref{Fig:teaser},~\ref{Fig:gallery1},~\ref{Fig:gallery2}) for qualitative assessment. All the code/data in this paper will be available to public for research purpose.

\subsection{Dataset}
We use a synthetic dataset to supervise the training for 1) correspondence detection, 2) shading prediction, and 3) UV gradient prediction. Our training dataset, named as \textsf{dataset-train}, is simple and not restricted to any specific type of garment. To do so, we randomly drop a piece of cloth onto a primitive shape (sphere, corn, etc.) in a physically based simulator (i.e., \cite{houdini}), and render the simulation sequence using Blender~\cite{blender}, with one of our generated patterns under an arbitrary illumination sampled from~\cite{envmap}. As shown in our supplementary material, our training set consists of 480 images at resolution $1K \times 1K$ from several simulation sequences rendered with 28 illuminations at random viewpoints. During the training, we randomly sample $30K$ patches from the rendered images at a resolution $96 \times 96$. As our dataset does not rely on a specific type of clothes, this guarantees the learned network can be generalized to arbitrary clothes during the application.

We also generate a separate synthetic dataset for quantitative evaluation. We collect several garments from \cite{santesteban2021self,santesteban2019virtualtryon,wang2019learning}. For each 3D garment, we randomly sample an environment illumination from ~\cite{envmap} and render videos from the front view. we generate the synthetic videos, \textsf{dataset-eval}, by running physically based garment simulation~\cite{houdini} on human animation sequences collected from~\cite{mikumiku,mahmood2019amass}. The synthetic sequences have total $1200$ frames. Finally, we capture the videos of the fabricated garment dressed by real human models as \textsf{dataset-real} to evaluate the quantitative and qualitative result of our method.

\subsection{Implementation}
\noindent\textbf{Pattern collection} 
Since multiple clothes may appear at the same time for our application, we need more area on the pattern space to sew the garment. In our practice, we use the pattern resolution of $100 \times 100$. The pattern searching method is the same as~\cite{halimi2022garment}. The pattern searching is fast and it takes no more than one second to generate a unique pattern. Although a $3\times3$ neighbor may not be unique in our pattern collection, the multi pattern voting strategy \textit{w.r.t.} the neighboring $3\times3$ patches makes sure the right pattern is identified. For multiple garments, we also add offsets to the UV locations of correspondences of different patterns to prevent overlapping of UV space in different garments.

\noindent\textbf{Pattern fabrication} The designed pattern is then physically fabricated by printing over a piece of cloths. We scale a single grid in our pattern to $15mm$ during the fabrication so a pattern of $100\times 100$ grids gives us a $1.5m \times 1.5m$ square cloths. The most commonly used digital fabric printing technology~\cite{DigitalFabric} is cost-effective and highly accurate. It also applies to different types of cloth material, including cotton, linen, silk, etc., which offers wide range of application scenarios. We fabricate 7 different types of garments using the patterns from our collection, namely, \textsf{T-shirt}/\textsf{waistcoat}/\textsf{long dress}/\textsf{skirt}/\textsf{pants} in Figure~\ref{Fig:gallery2}, \textsf{layered skirt} in Figure~\ref{Fig:blur} and \textsf{shorts} in supplementary video. This covers the most common types of garments and demonstrates the wide range of application scenarios of our method.

\noindent\textbf{Training details} 
The function $f$ in Eq~\ref{eqn:funcf} and Eq~\ref{eqn:funcmodel} is implemented by a MLP of 4 layers, each layer has $256$ dimension and activated by a softplus function ($\beta = 1, \mathrm{threshold}=20)$. Random Fourier Feature~\cite{tancik2020fourfeat} is used to improve the learning ability, with $\sigma$ setting to $1$. In our experiments, we find $\lambda_{grad}=5\times10^{-3}, \lambda_{temp}=0.7$ works stably well for all the test cases. For each kind of loss term, we randomly sample 10K points from all the frames of the video as a batch. We train 90K iterations for each video, with learning rate decaying from $10^{-3}$ to $10^{-5}$. It takes about two hours to train a video of 300 frames on a NVIDIA 4090 GPU. Please refer to the supplementary material for the detailed time usage. For the sparse correspondence detection, we follow the same procedure proposed in GarmentAvatar~\cite{halimi2022garment}, except our improvements on multiple patterns and searching from graph. For shading and UV gradient training and prediction, we adopt the U-Net architecture from~\cite{ronneberger2015u}. The shading layer can be derived from \textsf{dataset-train} as a form of shading supervision. Additionally, the UV gradient can be generated as a gradient supervision by applying a Sobel kernel to the rendered UV map. We train both shading and UV gradient U-Net for about 375K iterations, with batch size setting to 32. It takes about 10 hours to finish the training on a single 3060ti GPU. We also conduct data augmentation, like motion blur, image compression, noise, brightness/contrast/saturation adjustment, rotation, etc., on image patches to make the synthetic data resemble the real ones.

\subsection{Baselines}
We compare our approach with potential alternative baselines. Starting from the detected correspondences, directly perform 1) bi-linear or 2) RBF interpolation over the image domain can be a simple way to recover the dense UV coordinates. However, it falls short around wrinkles and seams. We also include 3) HumanGPS~\cite{tan2021humangps}, 4) Deformable Sprites (DS)~\cite{ye2022deformable}, 5) DensePose~\cite{guler2018densepose} and 6) Jafarian et al.~\cite{jafarian2023normal} as our baselines. Note that to align the UV space predicted by different approaches, we perform Procrustes analysis by finding the optimal rigid transformation in UV space to minimize the difference for~\cite{tan2021humangps,ye2022deformable,guler2018densepose,jafarian2023normal}. As shown in Figure~\ref{Fig:comparison}, these methods fail to produce visually plausible UVs, especially the UVs around the seam areas (highlighted in the green boxes). Please see the supplementary video for more clear comparisons. Due to the lack of data or code, the comparison with~\cite{neverova2020continuous,ianina2022bodymap,halimi2022garment} is not conducted. 

\subsection{Evaluation Metric}
\noindent 1) \textbf{Correspondence detection error.} We first evaluate our correspondence detection module. On \textsf{dataset-eval}, we report the average precision (AP) and recall (AR) for the detection module in the floating table. 
Discussing the thresholds for true positive is less interesting as our correspondence detection and voting strategy makes the image space location and UV coordinates either very accurate (e.g., less then 1 pixel error in image space if detected, and less than 1mm error in UV domain if voted yes) or very far away. Instead, we report AP and AR \textit{w.r.t.} the coordinate range (\textrm{max} - \textrm{min}) of the groundtruth UV coordinates from the given patch, as shown in table. 
\begin{wraptable}{r}{5.5cm}
\resizebox{0.6\columnwidth}{!}{
\begin{tabular}{|ccc|}
\toprule 
UV range    & $\leq 100mm$    & $>100mm$          \\ \hline
Proportion         & 66.5\% & 33.5\%  \\
AP\textsubscript{Ours}         & 99.9\% & 99.6\%  \\
AP\textsubscript{\cite{halimi2022garment}}   & 99.9\% & 99.9\% \\
AR\textsubscript{Ours}          & 98.7\% & 87.4\% \\
AR\textsubscript{\cite{halimi2022garment}}       & 95.5\% & 68.4\%  \\
\bottomrule \end{tabular}}
\end{wraptable} 
Higher range ($>100mm$) implies significant changes in UV coordinates, usually cased by high distortion or discontinuity, which is more challenging for correspondence detection. The proportion of these patches in \textsf{dataset-eval} is also presented. Compared to GarmentAvatar~\cite{halimi2022garment}, we achieve around $20\%$ improvement on the averaged recall for the top $30\%$ challenging cases on this task with negligible decrease on the averaged precision. 

We also evaluate the reliability when multiple patterns are used at the same time. In our voting method, we use three overlapped $3 \times 3$ patches to identify which pattern these patches belong to. We would like to see if different patterns might share similar local structures which could mislead our correspondence search. For this purpose, 1) we randomly select two patterns, \textsf{A} and \textsf{B}, from our pattern collection each time. 2) We randomly select three overlapped $3\times 3$ patches \textsf{Ax}, \textsf{Ay} and \textsf{Az} from \textsf{A}, and 3) search for the same patches \textsf{Bx}, \textsf{By} and \textsf{Bz} in pattern \textsf{B}. We repeat 1)-3) for $100\mathrm{K}$ times. We find that for all the cases, at least one of \textsf{Bx}, \textsf{By} and \textsf{Bz} does not exist, which means no similar local structure could be found in any two patterns. Such results indicate our correspondence search algorithm is robust to multiple patterns, and can be safely used when multiple garments are presenting simultaneously.

\begin{table}[]
\caption{Quantitative evaluation on \textsf{dataset-eval}.}
\vspace{-3mm}
\resizebox{\columnwidth}{!}{
\begin{tabular}{c|cccc}
\toprule 
                       & MSE(mm) $\downarrow$    & PSNR $\uparrow$          & SSIM $\uparrow$           & LPIPS $\downarrow$           \\ \hline
Ours                   &  \textbf{16.82$\pm$8.58}  & \textbf{8.51$\pm$0.99}  & \textbf{0.710$\pm$0.068}  & \textbf{0.146$\pm$0.043}    \\
Linear interp.         &  75.04$\pm$61.20          & 8.15$\pm$0.94           & 0.657$\pm$0.058           & 0.399$\pm$0.137     \\
RBF interp.            &  54.39$\pm$40.71          & 8.25$\pm$0.90           & 0.684$\pm$0.061           & 0.221$\pm$0.063     \\
humanGPS               &  138.18$\pm$28.36          & 7.10$\pm$0.20           & 0.600$\pm$0.012           & 0.701$\pm$0.054     \\
DensePose              &  86.82$\pm$16.18          & 7.24$\pm$0.15           & 0.569$\pm$0.020           & 0.502$\pm$0.030     \\
DS                     &  70.85$\pm$3.94          & 7.16$\pm$0.14           & 0.595$\pm$0.016           & 0.394$\pm$0.034     \\
\bottomrule 
\end{tabular}}
\label{table:comparison}
\end{table}

\noindent 2) \textbf{Per-pixel UV error.} We quantitatively evaluate our UV prediction on \textsf{dataset-eval}. In Table~\ref{table:comparison}, We report the average mean-squared error (MSE) between the predicted and the ground-truth UV coordinates. We also compute peak signal-to-noise ratio (PSNR)~\cite{5596999}, the complex wavelet structural similarity (CW-SSIM)~\cite{5596999,sampat2009complex} and perceptual similarity (LPIPS)~\cite{zhang2018perceptual} by texturing the patch with the input pattern using the predicted UV coordinates and compute the error \textit{w.r.t.} the input patch. Using CW-SSIM instead of SSIM is to reduce the influence of small shifting~\cite{sampat2009complex}. We compare our approach with the baselines: bi-linear interpolation, RBF interpolation, HumanGPS~\cite{tan2021humangps}, DensePose~\cite{guler2018densepose} and DS~\cite{ye2022deformable}. In table~\ref{table:comparison}, we show that our approach outperforms the baselines by a wide margin. A visual comparison between our approach and the baselines is shown in Figure~\ref{Fig:comparison}.

\begin{figure}[]
  \begin{center}
    \includegraphics[width=0.490\textwidth]{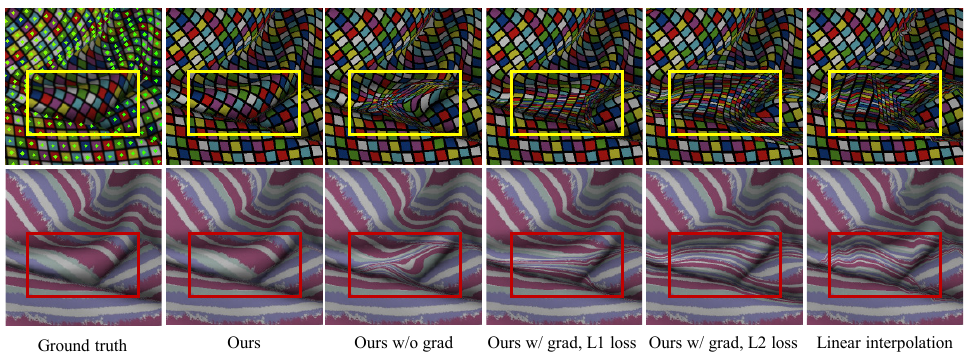}
  \end{center}
  \vspace{-5mm}
  \caption{The green dots represents the detected correspondences. The discontinuity in UV space may cause enormous artifacts for the interpolation-based approach. Our method trained with the gradient constraint and $L_{0.5}$ loss produces the best result.}
  \label{Fig:uv_discontinuity}
\end{figure}

\begin{table}[]
\caption{Quantitative evaluation on \textsf{dataset-real}.}
\vspace{-3mm}
\resizebox{\columnwidth}{!}{
\begin{tabular}{c|cccc}
\toprule
                & Ours & Ours\textsubscript{w/o temp} &Ours\textsubscript{w/o blend}  & DS  \\ \hline 
  tOF error $\downarrow$  & \textbf{1.70$\pm$0.62}  &  2.32$\pm$0.99   &  1.95$\pm$0.67 & 9.31$\pm$3.98   \\
  $\mathcal{L}_{consist}$ $(\times 10^{-2})$ $\downarrow$  & \textbf{1.10$\pm$0.80}  &  1.66$\pm$1.07   &  1.18$\pm$0.83 & 2.14$\pm$1.21   \\
  \bottomrule
\end{tabular}
}
\label{table:temporal}
\end{table}

\noindent 3) \textbf{Temporal consistency.} We evaluate the temporal coherency of our results from two aspects on \textsf{dataset-real}. 1) We texture the UV with the pattern, compute the optical flow on the textured video using~\cite{teed2020raft}, and compare the differences of optical flow between the original and textured video, which is defined as tOF error $E_{tOF} = \sum_t ||opt^t(\mathbf{V}_e)-opt^t(\mathbf{V}) ||\cdot S^t$, where $\mathbf{V}_e, \mathbf{V}$ is the textured video and original video, $opt^t$ means the optical flow at time $t$, and $S^t$ is the garment mask at time $t$. We also compare the $\mathcal{L}_{consist}$ term to evaluate the consistency. In Table~\ref{table:temporal}, we show our approach achieves better temporal consistency comparing to state-of-the-art video based decomposition approach DS~\cite{ye2022deformable}; 2) we show real captured video sequences for qualitative evaluation in our supplementary video. Please refer to our supplementary video for a perceptual comparison.

\begin{figure}[b]
  \begin{center}
    \includegraphics[width=0.475\textwidth]{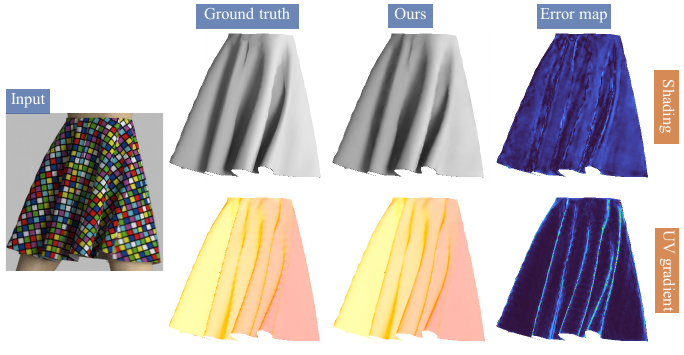}
  \end{center}
  \vspace{-5mm}
  \caption{An example of our shading/UV gradient prediction. }
  \label{Fig:shading}
\end{figure}
    
\noindent 4) \textbf{Shading/mask prediction error.} We predict the shading and UV gradient for the images in \textsf{dataset-eval} and compare them with the ground truth. Our shading prediction achieves $0.0228$ in $L_1$ loss, assuming the shading scale is $[0,1)$. For UV gradient prediction, we compare the $L_1$ loss of the vector length. We achieve the loss of value $0.08$, compared with the the average vector length $1.0$. We show a visual example for shading and UV gradient prediction in Figure~\ref{Fig:shading}, which proves our approach produces highly accurate shading and UV gradient from the input photo.

\noindent 5) \textbf{Model size.} The blended weight technique can greatly reduce the model size, compared with using a separate MLP for each frame of the video (Ours\textsubscript{w/o blend}). When $s$ setting to $10$ in Eq~\ref{eqn:cubicinter}, for a video of 270 frames, the model size of our method is 28MB while Ours\textsubscript{w/o blend} is 240MB. The model size of Ours\textsubscript{w/ blend} is about $12\%$ of the size of Ours\textsubscript{w/o blend}.  

\begin{figure}[b]
  \begin{center}
    \includegraphics[width=0.475\textwidth]{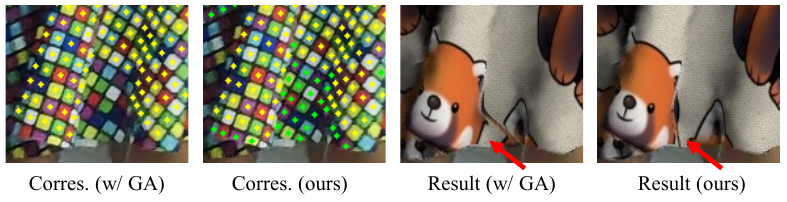}
  \end{center}
  \vspace{-5mm}
  \caption{Comparison between Ours and Ours\textsubscript{w/ GA}. Ours\textsubscript{w/ GA} cannot detect enough correspondences and produces the inaccurate UV. Our method increases the detected correspondence number and obtains better visual result.}
  \label{Fig:corres_improve}
\end{figure}

\begin{table}[]
\caption{Ablation study on \textsf{dataset-eval}.}
\vspace{-3mm}
\resizebox{\columnwidth}{!}{
\begin{tabular}{c|cccc}
\toprule 
                       & MSE(mm) $\downarrow$    & PSNR $\uparrow$          & SSIM $\uparrow$           & LPIPS $\downarrow$           \\ \hline
Ours                   &  \textbf{16.82$\pm$8.58}  & \textbf{8.51$\pm$0.99}  & \textbf{0.710$\pm$0.068}  & \textbf{0.146$\pm$0.043}    \\
Ours\textsubscript{w/ GA}  &    18.88$\pm$11.33    &     8.44$\pm$0.96    &    0.704$\pm$0.065    &   0.152$\pm$0.041  \\
Ours\textsubscript{w/o grad} &    19.79$\pm$9.63   &      8.36$\pm$0.86   &    0.695$\pm$0.059    &   0.201$\pm$0.050    \\ 
Ours\textsubscript{w/o blend} &   17.49$\pm$9.37   &      8.46$\pm$0.92   &    0.707$\pm$0.065    &   0.149$\pm$0.043    \\ 
Ours\textsubscript{w/o temp} &    36.04$\pm$24.48   &      8.43$\pm$1.03   &    0.700$\pm$0.070    &   0.153$\pm$0.051    \\ 
Ours\textsubscript{w/o temp, w/o blend} &   42.55$\pm$32.18   &      8.40$\pm$1.06   &    0.697$\pm$0.070    &   0.166$\pm$0.068    \\ 
\bottomrule 
\end{tabular}}
\label{table:ablation}
\end{table}

\subsection{Ablation Study}

\begin{table}[]
\caption{Quantitative evaluation of different model architectures on \textsf{dataset-eval}.}
\vspace{-3mm}
\resizebox{\columnwidth}{!}{
\begin{tabular}{c|cccccc}
\toprule 
    & MSE(mm) $\downarrow$    & PSNR $\uparrow$          & SSIM $\uparrow$           & LPIPS $\downarrow$  & $\mathcal{L}_{data}$ $\downarrow$  & model size              \\ \hline
$s=1$  &    17.09$\pm$9.09    &     8.50$\pm$0.99    &    0.710$\pm$0.068    &   0.148$\pm$0.044  & \textbf{0.57e-3} & 300MB \\
$s=10$ &  \textbf{16.82$\pm$8.58}  & \textbf{8.51$\pm$0.99}  & \textbf{0.710$\pm$0.068}  & \textbf{0.146$\pm$0.043}  & 0.58e-3 & 32MB \\
$s=100$ &    18.20$\pm$8.32   &      8.25$\pm$0.77   &    0.693$\pm$0.054    &   0.198$\pm$0.043  & 1.88e-3  & 5.7MB\\ 
$f(\mathbf{x},t)$ &   24.96$\pm$11.53   &      7.84$\pm$0.62   &    0.658$\pm$0.042    &   0.249$\pm$0.055  & 4.54e-3 & 1.5MB\\ 
\bottomrule 
\end{tabular}}
\label{table:modelstruc}
\end{table}

We conduct an ablation study to analyze the impact of our design choices. In Table~\ref{table:ablation} (Ours\textsubscript{w/ GA}), we show that with our improved correspondence search algorithm contributes to the improvement on UV accuracy compared to using the correspondence searching algorithm from~\cite{halimi2022garment}. We also show the visual comparison in Figure~\ref{Fig:corres_improve}.

For the $L_{0.5}$ gradient loss described in Section~\ref{sec:dense}, we show that without the loss, the method creates strong artifacts as shown in Figure~\ref{Fig:uv_discontinuity}, as well as larger quantitative errors as reported in Table~\ref{table:ablation}. The $L_{0.5}$ loss is also essential for the pleasant result compared with $L_{1}$ and $L_{2}$ loss, as shown in Figure~\ref{Fig:uv_discontinuity}. We also evaluate the UV prediction error on the detected correspondence locations to assess the quality of the regression. The mean squared error on correspondence point for our approach is $0.93\pm0.58 mm$, and $0.92\pm0.54 mm$ for our approach trained without the gradient term. We see that the use of the gradient loss clearly improves the performance around edges quantitatively and qualitatively, while the affection to the regression task itself is negligible. We further conduct ablation study on the temporal loss and the blended weight technique. In Table~\ref{table:temporal}, we see that our temporal loss and the blended weight MLP is helpful in improving the result. Please refer to supplementary video for the qualitative evaluation of ablation study.

We also test our method with different configurations of the model architecture. We set $s=1,10,100$ to evaluate the impact of the choice of $s$. We also test with a single MLP model $\mathbf{u} = f(\mathbf{x}, t; \bm{\theta}) : \mathbb{R}^3 \mapsto \mathbb{R}^{2}$ for the whole video sequence, which takes time $t$ as input. As shown in Table~\ref{table:modelstruc}, $s=1,10$ obtain better qualitative results. With the same training settings, $s=1,10$ could converge better and have lower data loss, which proves better learning ability of the models. We also show the model size when training on a video of 300 frames in Table~\ref{table:modelstruc}. Since the model size is quite large when setting $s=1$ , we choose $s=10$ in our applications for the balance of the model size and the model learning ability.

\begin{figure}[t]
  \begin{center}
    \includegraphics[width=0.475\textwidth]{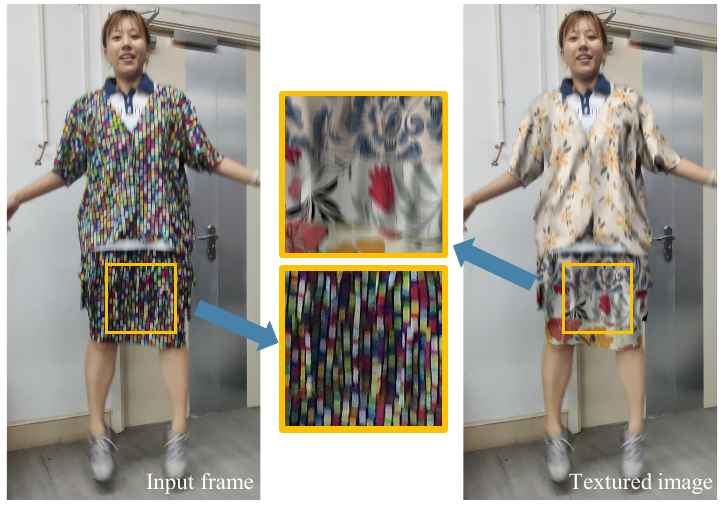}
  \end{center}
  \vspace{-5mm}
  \caption{Blurred image example. Our method is robust to severe motion blur. Please note that we extract the blurry kernel from the image and apply it back to our output textured image to restore the dynamic effects.}
  \label{Fig:blur}
\end{figure}

\begin{figure}[t]
  \begin{center}
    \includegraphics[width=0.475\textwidth]{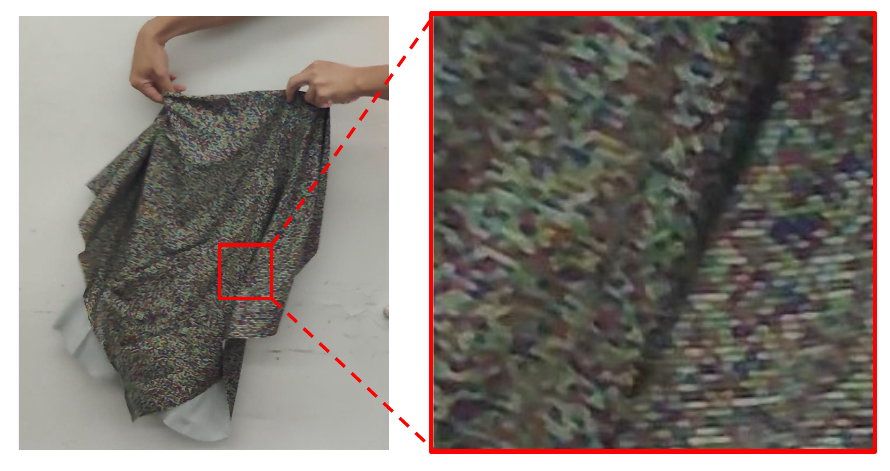}
  \end{center}
  \vspace{-5mm}
  \caption{A failure case of high pattern resolution. We experiment with the higher pattern resolution of 2.7mm, which is the same as GarmentAvatar\cite{halimi2022garment}. We capture the video with a hand-hold phone camera in the indoor illumination. Other settings are the same as cases in Figure~\ref{Fig:gallery2}. The image is so blurry that no correspondence could be detected.}
  \label{Fig:3mm_failure}
\end{figure}

\begin{figure}[t]
  \begin{center}
    \includegraphics[width=0.475\textwidth]{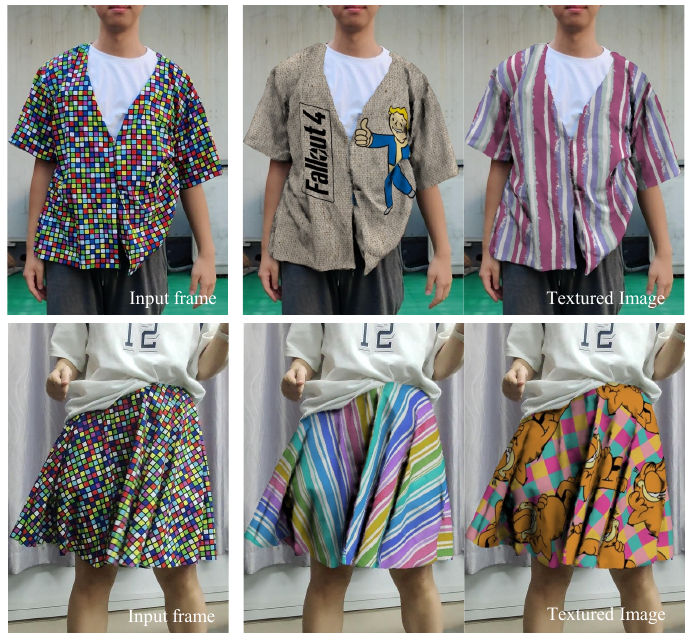}
  \end{center}
  \vspace{-5mm}
  \caption{Input frame with different textures.}
  \label{Fig:gallery1}
\end{figure}

\begin{figure*}[p]
  \begin{center}
    \includegraphics[width=0.990\textwidth]{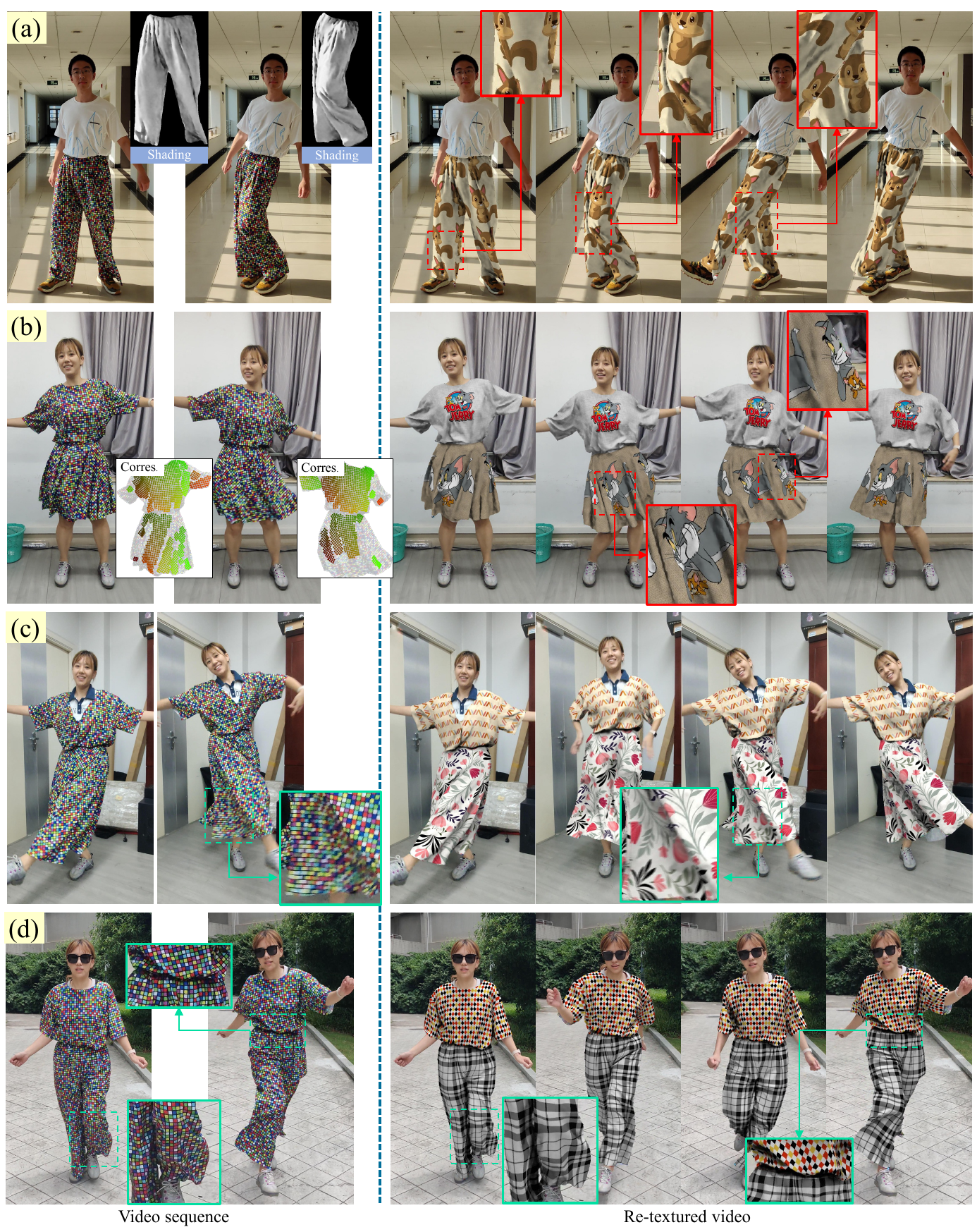}
  \end{center}
  \vspace{-5mm}
  \caption{Results on videos. Our approach is robust and applicable to a wild range of scenarios, including indoor and outdoor scenes, challenging motions, various illuminations, etc..}
  \label{Fig:gallery2}
\end{figure*}

\subsection{Application}

We record video footage for our garments and use them to test our approach. We note that all of the real videos used in the paper are captured by an ordinary phone camera and \textbf{NO} special device and setting is required. The captured video has 30 fps, with a resolution of $1920\times1080$. Here we show some examples in Figure~\ref{Fig:blur}, Figure~\ref{Fig:gallery1} and Figure~\ref{Fig:gallery2}. Note that our method is model-agnostic so is applicable to full body or partial body (Figure~\ref{Fig:gallery1}), single or multiple patterns (Figure~\ref{Fig:gallery2}, a and b). It is applicable to garment with sharp wrinkles (Figure~\ref{Fig:gallery1}, b) and challenging body pose and dynamics, like dancing and jumping (Figure~\ref{Fig:blur}, Figure~\ref{Fig:gallery2}, c, d). Our method can work at different scenarios, including indoor and outdoor scenes with various illuminations (Figure~\ref{Fig:gallery2}). Our method is capable of working with a high resolution (1080p) for better quality. Please refer to our supplementary video for more visual results. 

\subsection{Limitation and Future Work}

Our method cannot recover the UV from small objects or decorations, like the bowknot or the belt. Increasing the pattern resolution may work, but it may demand a camera of higher quality and sacrifice the robustness of the method. Figure~\ref{Fig:3mm_failure} shows a failure case when using the higher pattern resolution as GarmentAvatar~\cite{halimi2022garment}. Also, our results still look somewhat synthetic compared to the real captured data. We plan to model the specularity and transparency of the fabric material and use physical based rendering to further enhance the realism. We will also seek a better mask prediction method to make the fabric surrounding the neck more stable.


\section{Conclusion}
\label{sec:conclusion}

We present an approach for texture replacement of fashion videos. Powered by a neural regression step which is capable of capturing the discontinuity and retaining consistency in UV domain, our solution offers plausible results and opens a new opportunity for advertising, online retailing and customization of fashion product.

\section*{Acknowledgments}
The authors would like to thank reviewers for their insightful comments, Beijia Chen for presenting in the videos and Siyuan Shen for video capturing.
This work was supported by the National Key Research and Development Program of China (No.2022YFF0902302) and the NSF China (No. 62322209 and No. U23A20311).


\bibliographystyle{IEEEtran}
\bibliography{TVCG-2023-09-0587_Bib}

\begin{IEEEbiography}[{\includegraphics[width=1in,height=1.25in,clip,keepaspectratio]{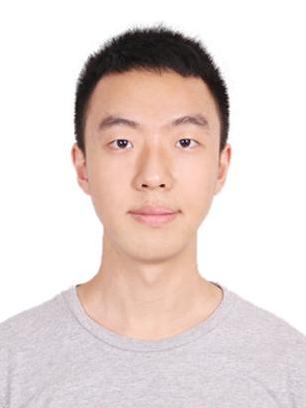}}]{Youyi Zhan}
Youyi Zhan is working toward the Ph.D. degree at the State Key Lab of CAD\&CG, Zhejiang University. His research interests include deep learning, image processing and garment animation.
\end{IEEEbiography}

\begin{IEEEbiography}[{\includegraphics[width=1in,height=1.25in,clip,keepaspectratio]{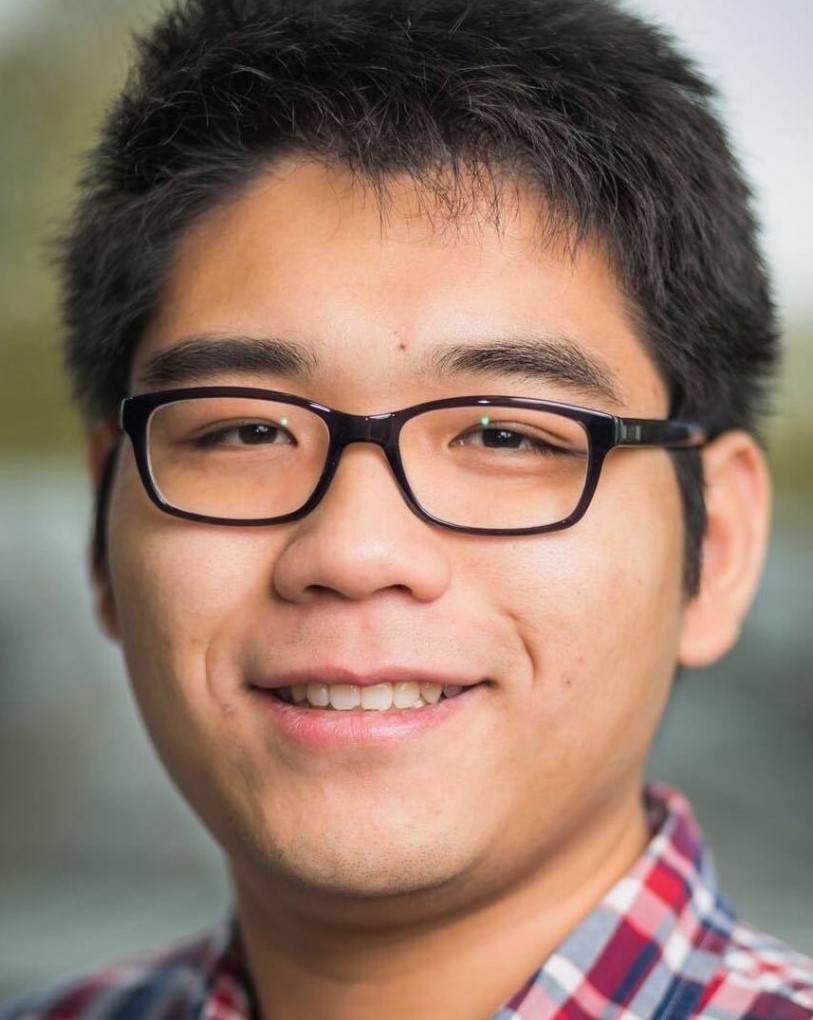}}]{Tuanfeng Y. Wang}
Tuanfeng Y. Wang is a Research Scientist at Adobe Research. He received his PhD from University College London, advised by Niloy J. Mitra, and B.S. from University of Science and Technology of China, advised by Ligang Liu. Tuanfeng's research interests are computer graphics and computer vision, including neural rendering, geometry processing, computational fabrication, computational geometry, 3D reconstruction, augmented/visual reality, etc.
\end{IEEEbiography}

\begin{IEEEbiography}[{\includegraphics[width=1in,height=1.25in,clip,keepaspectratio]{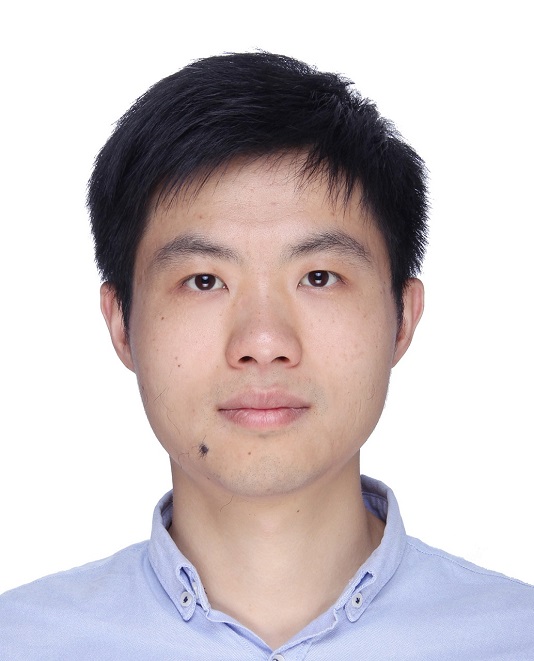}}]{Tianjia Shao}

Tianjia Shao is a Professor in the State Key Laboratory of CAD\&CG, Zhejiang University. Previously, he was an Assistant Professor in the School of Computing, University of Leeds, UK. He received his Ph.D. in Computer Science from Institute for Advanced Study, Tsinghua University, and his B.S. from the Department of Automation, Tsinghua University. His research interests include 3D scene/object modeling, digital avatar creation, structure/function aware geometry processing, computer animation, and 3D printing.
\end{IEEEbiography}

\begin{IEEEbiography}[{\includegraphics[width=1in,height=1.25in,clip,keepaspectratio]{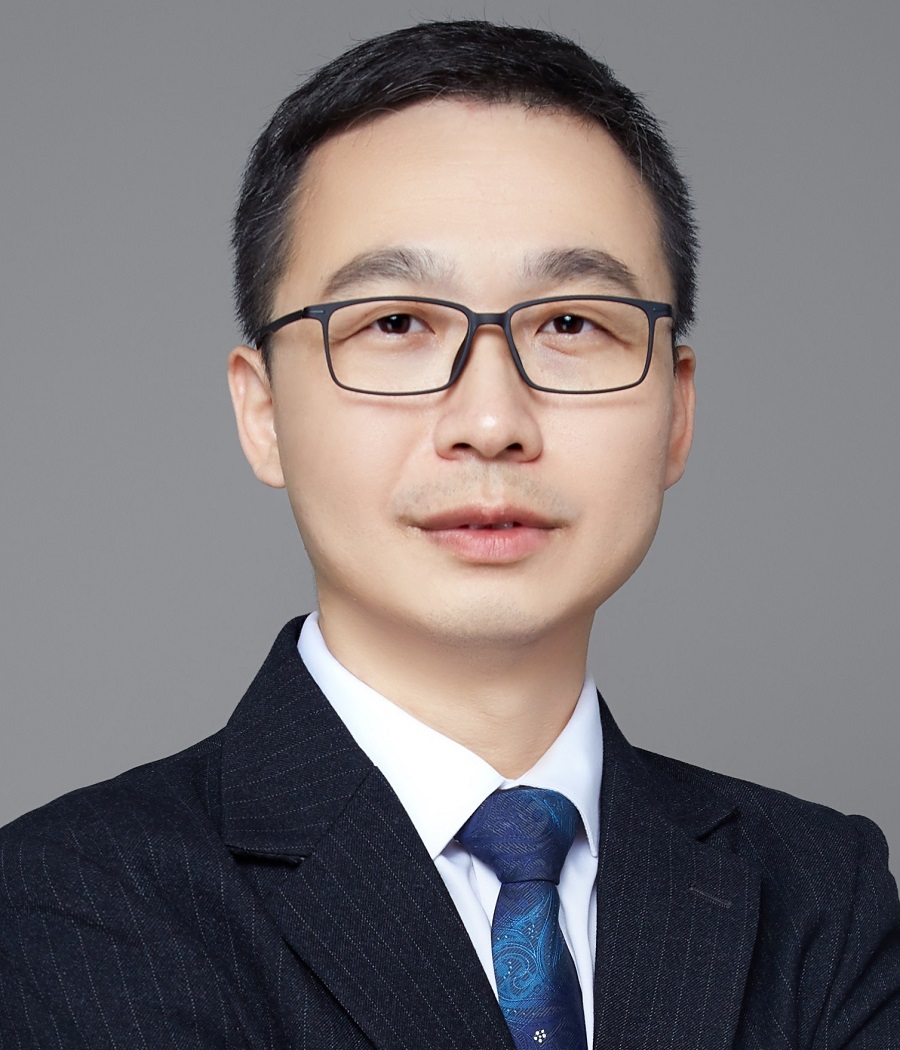}}]{Kun Zhou}
Kun Zhou is a Cheung Kong Professor in the Computer Science Department of Zhejiang University, and the Director of the State Key Lab of CAD\&CG. Prior to joining Zhejiang University in 2008, Dr. Zhou was a Leader Researcher of the Internet Graphics Group at Microsoft Research Asia. He received his B.S. degree and Ph.D. degree in computer science from Zhejiang University in 1997 and 2002, respectively. His research interests are in visual computing, parallel computing, human computer interaction, and virtual reality. He is a fellow of IEEE and ACM.
\end{IEEEbiography}

\clearpage





\setcounter{section}{0}  

\section{Negative Societal Impact}
Our garment texturing approach provides an efficient way for users to replace the texture of a garment if the video capture is given. This could lead to the generation of videos whose garment appearances do not exist in the real world. The users can potentially misuse this technology to create unreasonable marketing ads, thus impacting the society at large in a negative manner.

\section{Samples of \textsf{dataset-train}}
\begin{figure}[h]
  \begin{center}
    \includegraphics[width=0.475\textwidth]{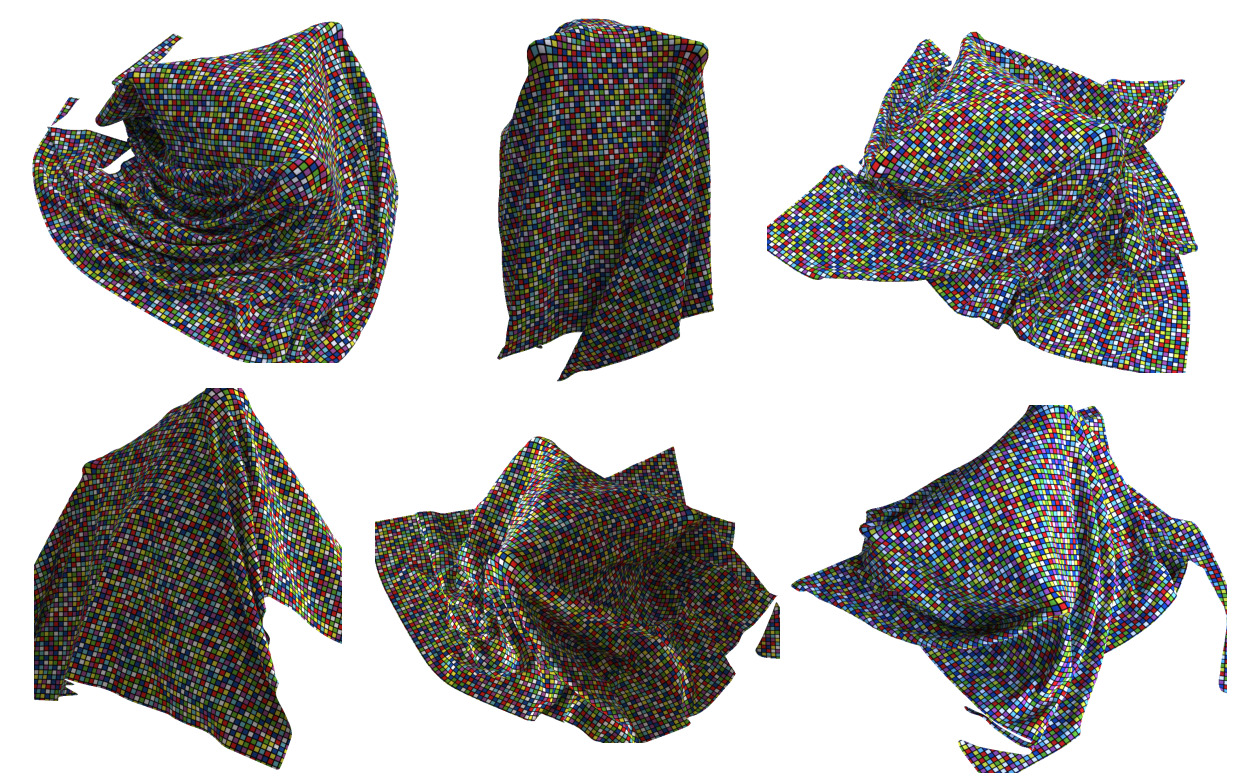}
  \end{center}
  \caption{Samples from \textsf{dataset-train}.}
  \label{fig:dataset}
\end{figure}

\begin{figure}[h]
  \begin{center}
    \includegraphics[width=0.475\textwidth]{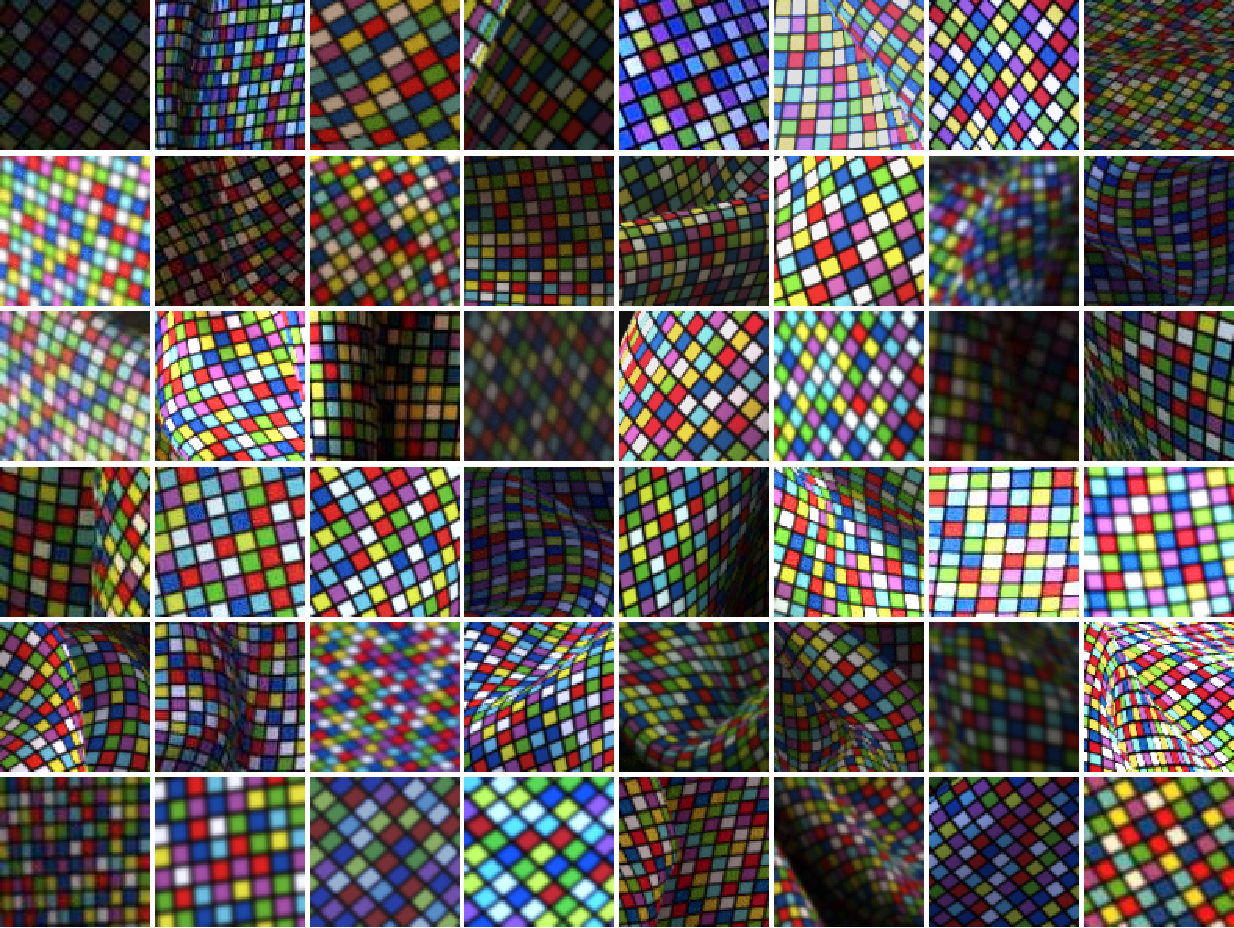}
  \end{center}
  \caption{Patches cropped from \textsf{dataset-train} for training correspondence detection, shading layer prediction and UV gradient.}
  \label{fig:dataset2}
\end{figure}

\begin{figure}[h]
  \begin{center}
    \includegraphics[width=0.475\textwidth]{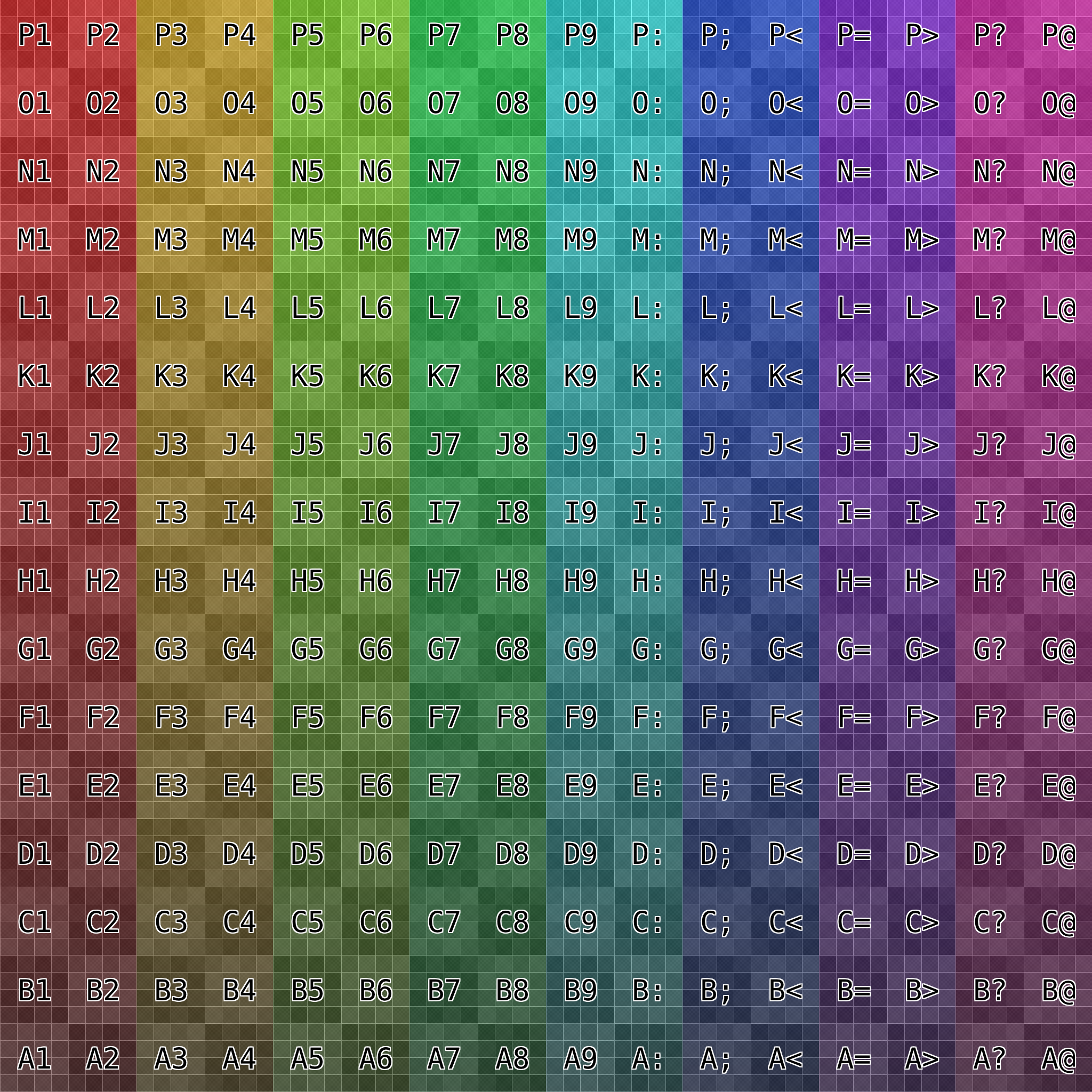}
  \end{center}
  \caption{The texture map used for our pipeline in Figure 3.}
  \label{fig:blender_texture}
  \vspace{-3mm}
\end{figure}

We show samples of \textsf{dataset-train} in Figure~\ref{fig:dataset}. Our dataset is not garment-type specific, so it can be generalized to almost all types of clothes during the test time. In Figure~\ref{fig:dataset2}, we further show some examples of the patches cropped for training the network for correspondence detection, UV gradient  and shading layer prediction. Figure~\ref{fig:blender_texture} gives the texture map we used for Figure 3 in the paper.

\section{Time of Workflow Components}
Here we list the processing time of different workflow components in Table~\ref{table:moduletime} on a 300 frame video using a single 4090 GPU. "First train" is used for obtaining UV gradient direction, as illustrated in Section 3.4, Paragraph 2 in the main paper, and "Second train" is for final video UV.

\begin{table}[h]
\caption{Time of workflow components}
\vspace{-3mm}
\resizebox{\columnwidth}{!}{
\begin{tabular}{c|ccccccc}
\toprule 
Total & Corres. detect  & Mask   & Optical flow  &  Shading & First train & UV gradient & Second train         \\ \hline
136.9m & 241s & 56s & 112s & 30s & 24.3m & 53s & 104.4m\\
\bottomrule 
\end{tabular}
}
\label{table:moduletime}
\end{table}

\section{UV Gradient Computation}
To extract $\mathbf{g}$ from the image, we use a U-Net to predict $\mathbf{g}$. The U-Net accepts inputs with seven channels: three for RGB image $\mathbf{I} \in \mathbb{R}^{H\times W \times 3}$, and four for the UV gradient direction $\mathbf{d} \in \mathbb{R}^{H\times W \times 2 \times 2}$. The output of the U-Net is a four channel array $\mathbf{g} \in \mathbb{R}^{H\times W \times 2 \times 2}$.

We now give the definitions of the UV gradient $\mathbf{g}$ and UV gradient direction $\mathbf{d}$. The UV gradient $\mathbf{g}$ is the convolution result of the UV map with Sobel kernel. In mathematics, $\mathbf{U},\mathbf{V}\in \mathbb{R}^{H\times W}$ is the U and V channel of a UV map. If we define the horizontal and vertical kernel as
\begin{equation}
\mathbf{K}_x = 
\begin{bmatrix} 
-1 & 0 & 1 \\
-2 & 0 & 2 \\
-1 & 0 & 1
\end{bmatrix}
\quad
\mathbf{K}_y = 
\begin{bmatrix} 
-1 & -2 & -1 \\
0 & 0 & 0 \\
1 & 2 & 1
\end{bmatrix},
\end{equation}
the UV gradient $\mathbf{g}$ can be computed as follows
\begin{equation}
\begin{aligned}
\mathbf{g}_{ux} &=\mathbf{K}_x * \mathbf{U} \quad \mathbf{g}_{uy} = \mathbf{K}_y * \mathbf{U} \\
\mathbf{g}_{vx} &=\mathbf{K}_x * \mathbf{V} \quad \mathbf{g}_{vy} = \mathbf{K}_y * \mathbf{V} \\
\mathbf{g} &= \{\{\mathbf{g}_{ux}, \mathbf{g}_{uy}\},\{ \mathbf{g}_{vx}, \mathbf{g}_{vy}\}\},
\end{aligned}
\end{equation}
where $*$ indicates the convolution operation, $\mathbf{g}_{ux}, \mathbf{g}_{uy}, \mathbf{g}_{vx}, \mathbf{g}_{vy} \in \mathbb{R}^{H\times W}$, $\mathbf{g} \in \mathbb{R}^{H\times W \times 2 \times 2}$. Note we clip the gradient value to suppress the large gradient value caused by the folding and seams. The UV gradient direction $\mathbf{d} \in \mathbb{R}^{H \times W \times 2 \times 2}$ is the UV gradient whose magnitude is normalized to $1$
\begin{equation}
\begin{aligned}
\mathbf{d}_u &= \{\mathbf{g}_{ux},\mathbf{g}_{uy}\}/\sqrt{\mathbf{g}_{ux}^2+\mathbf{g}_{uy}^2} \\
\mathbf{d}_v &= \{\mathbf{g}_{vx},\mathbf{g}_{vy}\}/\sqrt{\mathbf{g}_{vx}^2+\mathbf{g}_{vy}^2} \\
\mathbf{d} &=   \{ \mathbf{d}_u, \mathbf{d}_v \}.
\end{aligned}
\end{equation}
During training, $\mathbf{g}$ and $\mathbf{d}$ can be computed from groundtruth UV using above definition. During inference, $\mathbf{d}$ can be computed from Section 3.4 in the main paper. Overall, the gradient direction provides a rough direction guess of the actual gradient, which helps to eliminate the prediction ambiguity on the grid-like pattern.

\section{Correspondence Detection Algorithm}

We follow the correspondence detection method in GarmentAvatar~\cite{halimi2022garment}. The method contains 3 main modules, including 1) point and color detection, 2) the generation of heterogeneous graph and homogeneous graph, and 3) board location identification and neighbor voting. Based on this method, we propose an extra module 4) searching from the graph boundary, to further enhance the number of detected correspondences by growing the homogeneous graph boundary step by step, as mentioned in Sec 3.1 in the paper. We give a brief introduction of the first three modules in Section~\ref{sec:ori_corres}, and explain the searching method in Section~\ref{sec:search}. Please refer to GarmentAvatar~\cite{halimi2022garment} for more details of the first three modules.

\subsection{Original Correspondence Detection}
\label{sec:ori_corres}
\textbf{Point and Color Detection.} The method in GarmentAvatar~\cite{halimi2022garment} first applies a UNet to detect the center points and corner points of an input image. It also performs color detection with another UNet to predict the color of each pixel. Different from GarmentAvatar~\cite{halimi2022garment}, we use synthetic data to train the above networks, so that the size of the dataset can be much larger and the point and color labels are more accurate. We also apply data augmentation to adapt the network to the real world data.

\noindent\textbf{Generation of Heterogeneous Graph and Homogeneous Graph.} The heterogeneous graph is a graph in which an edge connects a center point and a corner point. The heterogeneous graph is used to eliminate the ambiguity of homogeneous graph construction under a general affine transformation, as mentioned in GarmentAvatar~\cite{halimi2022garment}. For the homogeneous graph, each of its edges connects two center points. It is used for the 3×3 grid extraction and location identification. We follow the same way as GarmentAvatar~\cite{halimi2022garment} to establish the two graphs. 

\noindent\textbf{Board Location Identification and Neighbor Voting.} Board location identification aims to assign a UV location on the pattern board to a detected center point on the image. Neighbor voting aims to eliminate potential identification errors. We basically follow these two steps. However, a key difference between our method and GarmentAvatar~\cite{halimi2022garment} is, our applications require multiple patterns for the different garments. To address this issue, we modify the neighbor voting algorithm to make it robust to multiple patterns. Specifically, for a $3\times 3$ grid, instead of doing neighbor voting on a single pattern, we do neighbor voting in all the used patterns and get their major votes. We choose the pattern with the highest vote count as the grid's correct pattern, and assign the corresponding pattern id and board location to the grid vertices. Please refer to Algorithm~\ref{algo:nervote} for the detailed process.

\subsection{Searching from The Graph Boundary}
\label{sec:search}

The original correspondence detection pipeline highly depends on the accuracy of heterogeneous graph and homogeneous graph. Even a single error in these graphs results in a small area undetected. However, at the highly skewed boundary areas (e.g., wrinkles and occlusions) or blurry areas of the image, the establishment of heterogeneous graph can fail. We make a key observation that despite such highly skewed areas, the center points and their color can still be correctly recognized there. We can design an efficient method to search more center points from the detected graph. Briefly speaking, at the detected homogeneous graph, we search for extra center points not existing in the homogeneous graph to construct a new $3\times 3$ grid and get the 9-color-bit code. If the hash-code inference result of the 9-color-bit code exists in a pattern and the location of inference result is compatible with the grid location, we accept the newly searched center points as new correspondence points. 

We use Figure 4 (top row) in the main paper as an example to explain the process, and the searching algorithm is given in Algorithm~\ref{algo:corrsearch}. Let $p_1$ be a center point near the graph boundary with a $3\times 3$ grid. $p_1$'s right neighbor is $p_3$, and $p_3$ doesn't have a $3\times 3$ grid around. Hence $p_3$ is at the homogeneous graph boundary. We define the up and down neighbors of $p_3$ are $p_2$ and $p_4$. Note $p_2$ and $p_4$ must exist, as $p_1$ has a full $3\times 3$ grid. We also define the right neighbors of $p_2, p_3, p_4$ as $p'_2, p'_3, p'_4$. Because $p_3$ doesn't have a full $3\times 3$ grid, at least one of $p'_2, p'_3, p'_4$ doesn't exist in the homogeneous graph. Our aim is to search for $p'_2, p'_3, p'_4$ to build a full $3\times 3$ grid around $p_3$. 
Considering the searching of $p'_3$ for example, $p'_3$ may appear on the extension of line $p_1p_3$, and $||p'_3-p_3||$ may almost equal to $||p_3-p_1||$. Therefore, we search for a center point $p'_3$ around $p''_3 = p_3 + \overrightarrow{p_3-p_1}$ with a threshold $\frac{1}{2}||p_1-p_3||$. If such center point $p'_3$ doesn't exist, the searching of $3\times 3$ grid around $p_3$ fails. The same searching process is also conducted for $p'_2, p'_4$.

If $p'_2, p'_3, p'_4$ are all found, $p_3$ now has a full $3\times 3$ grid around. We now check whether the search is correct or not. We extract the 9-color-bit code of the $3\times 3$ grid, and do hash code inference on pattern $T$ to get the location of the grid on the pattern. If the 9-color-bit hash code doesn't exist, or if the grid's UV location is inconsistent with point $p_3$'s location, we'll reject the grid and the searching of $3\times 3$ grid around $p_3$ fails.

If the above process succeeds, we can claim we have found new correspondences $p'_2, p'_3, p'_4$ and safely add them to the homogeneous graph. New edges related to the new point $p'_2, p'_3, p'_4$ are added to the homogeneous graph as well. We repeat the above search process for new center points. Once the graph boundary points reach the edge of a piece of cloth or some areas where the colors are mistakenly recognized or the center points can't be detected, the searching process stops in these cases. 

For multiple patterns, the above algorithm also works, as the searching process stops at the empty results of hash code inference or inconsistent UV locations when the pattern and the grid are incompatible.

\begin{algorithm}[t]
\SetAlgoLined
\DontPrintSemicolon
\SetKw{Continue}{continue}
\SetKw{Break}{break}
\newcommand\mycommfont[1]{\footnotesize\sffamily\textcolor{blue}{#1}}
\SetCommentSty{mycommfont}
\KwIn{The homogeneous graph $G=(V, E)$ \newline
    Used patterns $\{T_i\}$
    }
\KwOut{The vertices' board location $B(V)$ and pattern id $P(V)$ 
    }
\;
$Nbr_{3 \times 3}(v)$, $3 \times 3$ grid vertices around $v$ in $G$\;
\tcp{A complete grid-graph has 9 vertices. If $3 \times 3$ grid-graph is incomplete, we allow $Nbr_{3 \times 3}(v)$ to have fewer than 9 vertices}\;
initialize $B$, $P$ with NULL\;

\For{vertex $v \in V$}{
    maxVoteCount $\leftarrow$ 0\;
    \For{pattern $T \in \{T_i\}$}{
        votes $\leftarrow$ $\{\}$ \;
        \For{vertex $u$ in $\{u | v \in Nbr_{3\times 3}(u)\}$ }{
            \If{$|Nbr_{3 \times 3}(u)| \ne 9$ }{
                \Continue
            }
            $c$ $\leftarrow$ 9-color-bit codes of $Nbr_{3 \times 3}(u)$ \;
            boardLocation\_u $\leftarrow$ HashFunction($c$, $T$)\;
            \If{boardLocation\_u is not empty}{
                \Continue
            }  
            \tcp{Calculate $v$'s location according to $u$'s location}
            boardLocation\_v $\leftarrow$ getNbrLocation($u$,$v$,boardLocation\_u, $T$)\;  
            votes.append(boardLocation\_v)\;
        }
        
        majorVote, majorVoteCount $\leftarrow$ CalcMajorVote(votes)\;
        \If{majorVoteCount $\ge$ 3 and \\ majorVoteCount $>$ maxVoteCount}{
            $B(v)$ $\leftarrow$ majorVote\;
            $P(v)$ $\leftarrow$ $T$ \;
            maxVoteCount = majorVoteCount
        }
    }
}
\tcp{Now $v \in \{v|B(v)\ne \text{NULL}\}$ is assigned with correct board location and pattern id, we assign location and id to other vertices in $Nbr_{3 \times 3}(v)$ as well}
\For{vertex $v \in \{v|B(v)\ne \text{NULL}\}$}{
    \For{vertex $u \in Nbr_{3 \times 3}(v)$}{
        \If{$B(u) = \text{NULL}$}{
            $P(u)$ $\leftarrow$ $P(v)$ \;
            $B(u)$ $\leftarrow$ getNbrLocation($v$,$u$,$B(v)$,$P(v)$)\;  
        }
    }
}
Remove $\{v | B(v)= \text{NULL}\}$ and related edges in $G$\;
\caption{Neighbor-Voting algorithm on multiple patterns}
\label{algo:nervote}
\end{algorithm}

\begin{algorithm}[t]
\SetAlgoLined
\DontPrintSemicolon
\SetKw{Continue}{continue}
\newcommand\mycommfont[1]{\footnotesize\sffamily\textcolor{blue}{#1}}
\SetCommentSty{mycommfont}
\KwIn{The homogeneous graph $G=(V, E)$ \newline
    The graph vertices’ board location $B(V)$ \newline
    Used patterns $\{T_i\}$ \newline
    Center vertices $\tilde{V}$ which haven't been assigned board locations ($B(\tilde{V})$=NULL) \newline
    \tcp{After Algorithm~\ref{algo:nervote}, $B(V) \ne \text{NULL}$. So $V \cap \tilde{V} = \varnothing $}
    }
\KwOut{Board location $B(\tilde{V})$ and pattern id $P(\tilde{V})$}
\;
$Q$, a queue storing vertices $v$ satisfying $| Nbr_{3 \times 3}(v) | = 9$\;
\;
\For{pattern $T$ in $\{T_i\}$}{
    $Q \leftarrow \{\}$ \;
    Append $ \{ v | \ | Nbr_{3 \times 3}(v) | = 9\} $ to $Q$\;
    \While{$Q$ is not empty}{
        $p_1 \leftarrow Q.pop()$ \;
        \For{vertex $p_3$ in $\{p | (p, p_1) \in E\}$}{
            \If{$ | Nbr_{3 \times 3}(p_3) | = 9 $}{
                \Continue
            }
            \tcp{$p_3$ doesn't have a complete $3 \times 3$ grid in $G$. So $p_3$ is at the $G$'s boundary. We search for extra vertices for a complete grid}
            Search $p'_2, p'_3, p'_4$ in $\tilde{V}$ w.r.t Figure 4 (top row) in the main paper\;
            \If{one of ${p'_2, p'_3, p'_4}$ does not exist}{
                \Continue
            }
            $ s \leftarrow Nbr_{3 \times 3}(p_3) \cup \{ p'_2, p'_3, p'_4 \}$ \;
            \tcp{Vertices in $s$ can compose a $3 \times 3$ grid centered at $p_3$ now. We then extract colors and verify the location}
            $c \leftarrow$ 9-color-bit codes of $s$ \;
            boardLocation\_p3 $\leftarrow$ HashFunction($c$, $T$) \;
            \If{boardLocation\_p3 is empty or \\ boardLocation\_p3 $\neq$ $B(p_3)$}{
                \Continue
            }
            \tcp{Now $p_3$ has a valid $3 \times 3$ grid $s$. We then add the new vertices and edges to $G$}
            Move  $p'_2, p'_3, p'_4$ from $\tilde{V}$ to $V$\;
            Assign locations to $B(p'_2), B(p'_3), B(p'_4)$ \;
            Add edges related to $p'_2, p'_3, p'_4$ to $E$\;
            $P(p'_2), P(p'_3), P(p'_4) \leftarrow T$ \;
            $Q$.append($p_3$) \;
        }
    }
}

\caption{Search correspondences from graph boundaries}
\label{algo:corrsearch}
\end{algorithm}

\end{document}